\documentclass[conference]{IEEEtran}
\usepackage{graphicx} % Required for inserting images

\IEEEoverridecommandlockouts
\usepackage{cite}
\usepackage{amsmath,amssymb,amsfonts}
\usepackage{algorithm, algorithmicx}
\usepackage[noend]{algpseudocode}
\usepackage{graphicx}
\usepackage{textcomp}
\usepackage{xcolor}
\usepackage{subcaption}
\usepackage{multirow}
\usepackage{listings}
\usepackage{enumitem}
\usepackage{url}
\usepackage{pifont}
\usepackage{subcaption}
\usepackage{array}
\usepackage{xspace}
\usepackage{bbm}
\usepackage{xspace}
\usepackage{tcolorbox}
\usepackage{color}
\usepackage{framed}
\definecolor{shadecolor}{rgb}{0.92,0.92,0.92}

\usepackage{listings}
\definecolor{codebrown}{rgb}{0.8,0.44,0.2}
\definecolor{codegray}{rgb}{0.5,0.5,0.5}
\definecolor{codepurple}{rgb}{0.58,0,0.82}
\definecolor{backcolour}{rgb}{1.0,1.0,1.0}
\lstdefinestyle{mystyle}{
    backgroundcolor=\color{backcolour},   
    commentstyle=\color{codebrown},
    keywordstyle=\color{magenta},
    numberstyle=\tiny\color{codegray},
    stringstyle=\color{codepurple},
    basicstyle=\ttfamily\footnotesize,
    breakatwhitespace=false,         
    breaklines=true,                 
    captionpos=b,                    
    keepspaces=true,                 
    numbers=left,                 
    showspaces=false,                
    showstringspaces=false,
    showtabs=false,                  
    tabsize=2,
    xleftmargin=0.025\textwidth,
    xrightmargin=0.\textwidth
}
\newcommand{\Approach}[1]{\texttt{R$^3$}}

\def\BibTeX{{\rm B\kern-.05em{\sc i\kern-.025em b}\kern-.08em
    T\kern-.1667em\lower.7ex\hbox{E}\kern-.125emX}}
\begin{document}

\title{\Approach{}: On-device Real-Time Deep
Reinforcement Learning for Autonomous Robotics}

\author{Zexin Li$^{1}$
\xspace\xspace\xspace    Aritra Samanta$^{1}$
\xspace\xspace\xspace    Yufei Li$^{1}$
\xspace\xspace\xspace    Andrea Soltoggio$^{2}$
\xspace\xspace\xspace    Hyoseung Kim$^{1}$
\xspace\xspace\xspace    Cong Liu$^{1}$\\
$^{1}$University of California, Riverside \xspace\xspace\xspace
$^{2}$Loughborough University\\
{\tt\small \{zli536, asama004, yli927, hyoseung, congl\}@ucr.edu, a.soltoggio@lboro.ac.uk} \\\
}

\maketitle

\newcommand{\zexin}[1]{\textcolor{blue}{(zexin: #1})}

\begin{abstract}
Autonomous robotic systems, like autonomous vehicles and robotic search and rescue, require efficient on-device training for continuous adaptation of Deep Reinforcement Learning (DRL) models in dynamic environments. This research is fundamentally motivated by the need to understand and address the challenges of on-device real-time DRL, which involves balancing timing and algorithm performance under memory constraints, as exposed through our extensive empirical studies. This intricate balance requires co-optimizing two pivotal parameters of DRL training -- batch size and replay buffer size. Configuring these parameters significantly affects timing and algorithm performance, while both (unfortunately) require substantial memory allocation to achieve near-optimal performance.

This paper presents \Approach{}, a holistic solution for managing timing, memory, and algorithm performance in on-device real-time DRL training. \Approach{} employs (i) a deadline-driven feedback loop with dynamic batch sizing for optimizing timing, (ii) efficient memory management to reduce memory footprint and allow larger replay buffer sizes, and (iii) a runtime coordinator guided by heuristic analysis and a runtime profiler for dynamically adjusting memory resource reservations. These components collaboratively tackle the trade-offs in on-device DRL training, improving timing and algorithm performance while minimizing the risk of out-of-memory (OOM) errors. 

We implemented and evaluated \Approach{} extensively across various DRL frameworks and benchmarks on three hardware platforms commonly adopted by autonomous robotic systems. Additionally, we integrate \Approach{}  with a popular realistic autonomous car simulator to demonstrate its real-world applicability. Evaluation results show that \Approach{} achieves efficacy across diverse platforms, ensuring consistent latency performance and timing predictability with minimal overhead. Moreover, \Approach{} showcases versatility by handling varied optimization goals and adapting to fluctuating systems scenarios. 
\end{abstract}

\section{Introduction}

Deep Reinforcement Learning (DRL) has emerged as a promising field, showing a pervasive influence in various real-world applications.~\cite{mnih2015human,van2016deep,bellemare2017distributional,schaul2015prioritized,schulman2017proximal,schulman2015trust,haarnoja2018soft} One such application is autonomous vehicles (AVs), where DRL models need to adapt to ever-changing road and traffic conditions continually~\cite{kahn2018self,peng2018deepmimic}.
The models must swiftly learn and retrain based on new data and evolving scenarios, maintaining their responsiveness and capability for immediate decision-making~\cite{few-shot-continual-learning,polynomial-continual}.
Similarly, in robotic rescue, DRL models enable robots to navigate hazardous environments, locate survivors, and deliver assistance~\cite{aggravi2021haptic,heintzman2021anticipatory}.
These models must adapt to rapidly shifting and unpredictable conditions, requiring efficient on-the-spot training and retraining to ensure timely decision-making. These examples underline the critical need for efficient on-device DRL training with timing constraints, integrating (re)training within runtime inference to timely adapt to dynamic and evolving environments.

\begin{figure}[!tbp]
\centering
\includegraphics[width=0.48\textwidth]{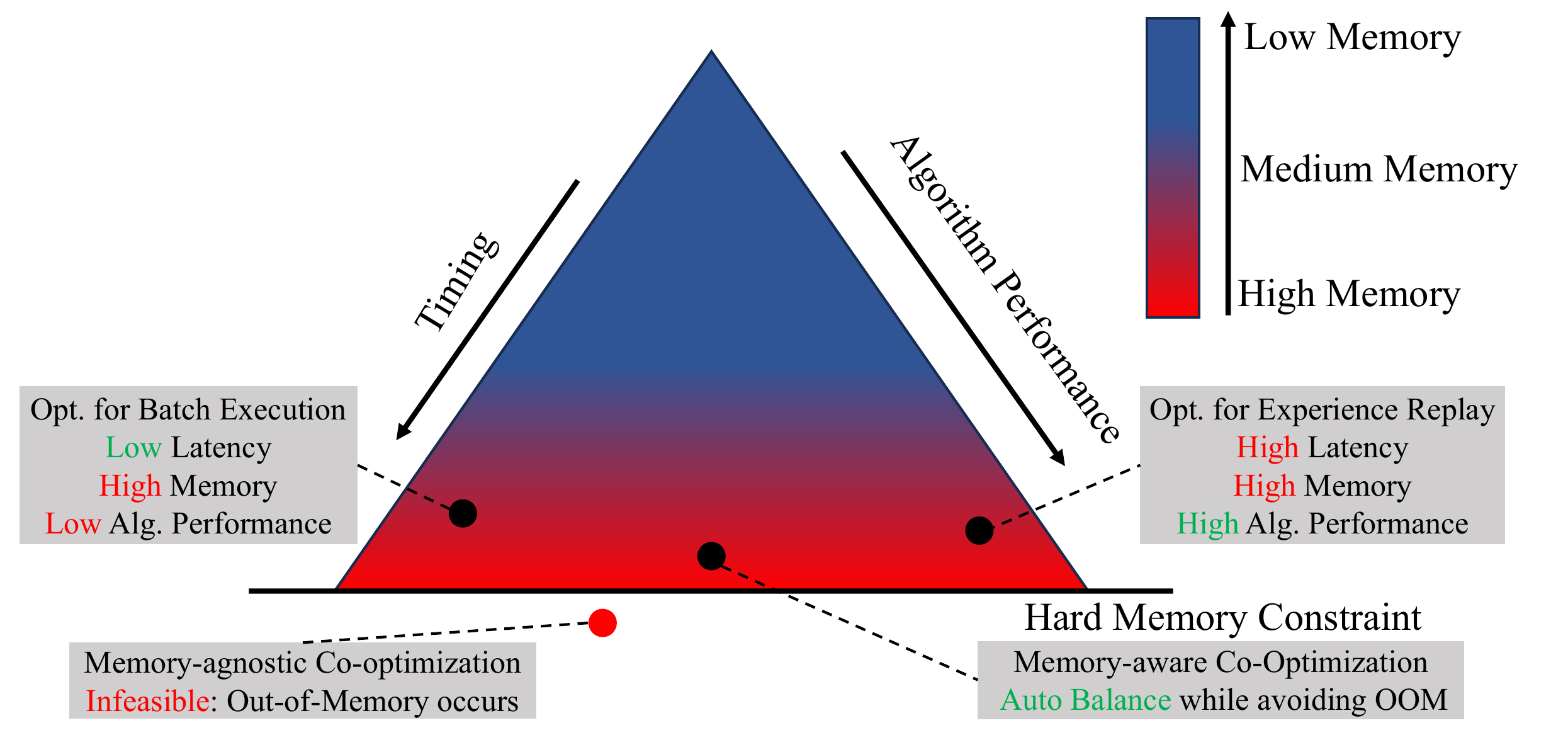}
\caption{Visualization of optimization challenges of embedded deep reinforcement learning training. Arrow direction means better timing, better algorithm performance, and less memory usage. A higher red level represents higher memory pressure.
}
\label{fig:R3_challenge_overview}
\vspace{-5mm}
\end{figure}

However, conducting on-device DRL training is fraught with unique challenges. As depicted by Figure~\ref{fig:R3_challenge_overview}, these challenges are fundamentally tied to the need for co-optimizing two frequently conflicting objectives: latency and algorithm performance\footnote{``Algorithm performance" in DRL is akin to ``accuracy" in DNNs. In DNNs, accuracy gauges the correct prediction rate. However, in DRL, performance is more multifaceted, primarily involving the agent's capability to optimize cumulative reward over time, balancing exploration and exploitation. While accuracy can be relevant in some tasks, DRL performance typically entails a broader, more complex set of considerations.}, a task made more complex by the memory constraints inherent to embedded devices. Intrinsically, the DRL training process involves two critical components: batch execution and experience replay, each representing its own dimension of trade-off.
Batch execution primarily pertains to the batch size parameter, which forms a crucial junction between timing and memory trade-offs. A larger batch size might accelerate the training process but increase memory consumption. On the other hand, experience replay entails a replay buffer size trade-off. Here, a larger buffer size can improve algorithm performance by providing more diverse experiences for training but can also cause memory issues due to the increased demand for storage.
Optimizing these trade-offs independently could result in less-than-ideal outcomes, while co-optimizing them in a memory-agnostic manner may lead to significant memory pressure or even trigger Out-of-Memory (OOM) errors which in some severe cases cause the memory-limited embedded system (e.g., 16 GB memory for AGX Xavier) to fail. Therefore, a holistic memory-aware approach that addresses both dimensions of trade-off simultaneously becomes indispensable. 
To navigate this intricate multi-dimensional trade-off space and manage the inherent complexity of \textbf{R}eal-time Deep \textbf{R}einforcement Learning for Autonomous \textbf{R}obotics, in this paper, we present \Approach{}, which is designed to tackle the complex problem of on-device DRL training with careful consideration of timing, memory, and algorithmic performance.
Our approach is composed of three innovative components. Firstly, we introduce a deadline-driven feedback loop, which dynamically assigns proper batch size for batch execution to balance memory and latency. Secondly, our approach includes efficient memory management, which applies task-specific memory optimizations to reduce memory footprint significantly. It allows for larger replay buffer sizes, enhancing algorithm performance in real-time embedded DRL while mitigating the risk of OOM occurrences. Finally, we incorporate a runtime coordinator, which synchronizes the interactions between the deadline-driven feedback loop, memory management, and other system components. This component, guided by heuristic analysis and a runtime profiler, dynamically adjusts memory resource reservations to meet hard resource constraints. These three components working together, enable \Approach{} to effectively balance the trade-offs of latency, memory, and algorithm performance in a holistic manner, ultimately achieving real-time on-device DRL training. 

Our approach, \Approach{}, has been deployed and tested across a range of DRL benchmarks~\cite{bib:gymdonkeycar,bellemare2013arcade,6313077} built upon different underlying frameworks~\cite{pytorch,tensorflow,nota2020autonomous,1606.01540} to assess its efficacy 
with its application extending to three platforms, including a desktop and two embedded devices \cite{bib:agx, bib:orin}. Moreover, we highlight the practical usability of \Approach{} by a further integration with a realistic autonomous car simulator \cite{bib:donkeycar_s1}. Empirical results indicate that \Approach{} achieves efficacy across diverse platforms, ensuring consistent latency performance and timing predictability with minimal overhead. Moreover, \Approach{} showcases versatility by handling varied optimization goals and adapting to fluctuating systems scenarios. Notable achievements of \Approach{} include:

\begin{itemize}[leftmargin=10px]
    \item \textbf{Cross-platform Efficacy:} Our approach has demonstrated its efficacy in diverse environments across different platforms, including Classic Control~\cite{6313077}, Atari~\cite{bellemare2013arcade}, and DonkeyCar~\cite{bib:gymdonkeycar} environments. (Sec.~\ref{sec:overall_effectiveness})
    \item \textbf{Latency predictability:} Our proposed method consistently achieves timing predictability and satisfies deadlines across all benchmarks. (Sec.~\ref{sec:overall_effectiveness})
    \item \textbf{Practical usability:} Our user-friendly approach can be seamlessly integrated into existing practical complex deep reinforcement learning systems, such as DonkeyCar~\cite{bib:gymdonkeycar}, without extensive modifications. (Sec.~\ref{sec:car})
    \item \textbf{Low overhead:} The design and implementation of our method are highly efficient, resulting in negligible overhead on execution time and memory. (Sec.~\ref{sec:overhead})
    \item \textbf{Versatility:}  Our proposed framework exhibits a high degree of versatility, adeptly addressing diverse optimization goals and adapting to varying system scenarios. (Sec.~\ref{sec:interference_adaptability})
\end{itemize}

\section{Background and Motivational Case Study}

\label{sec:case_study}

In this section, we present a series of illustrative case studies to elucidate the unique challenges associated with embedded deep reinforcement learning. These examples provide valuable insights into the limitations of existing approaches and highlight the potential pitfalls when naively extending them to address the problem context under consideration.

\subsection{Characteristics of Deep Reinforcement Learning}

\begin{figure}[!t]
\centering
\includegraphics[width=0.4\textwidth]{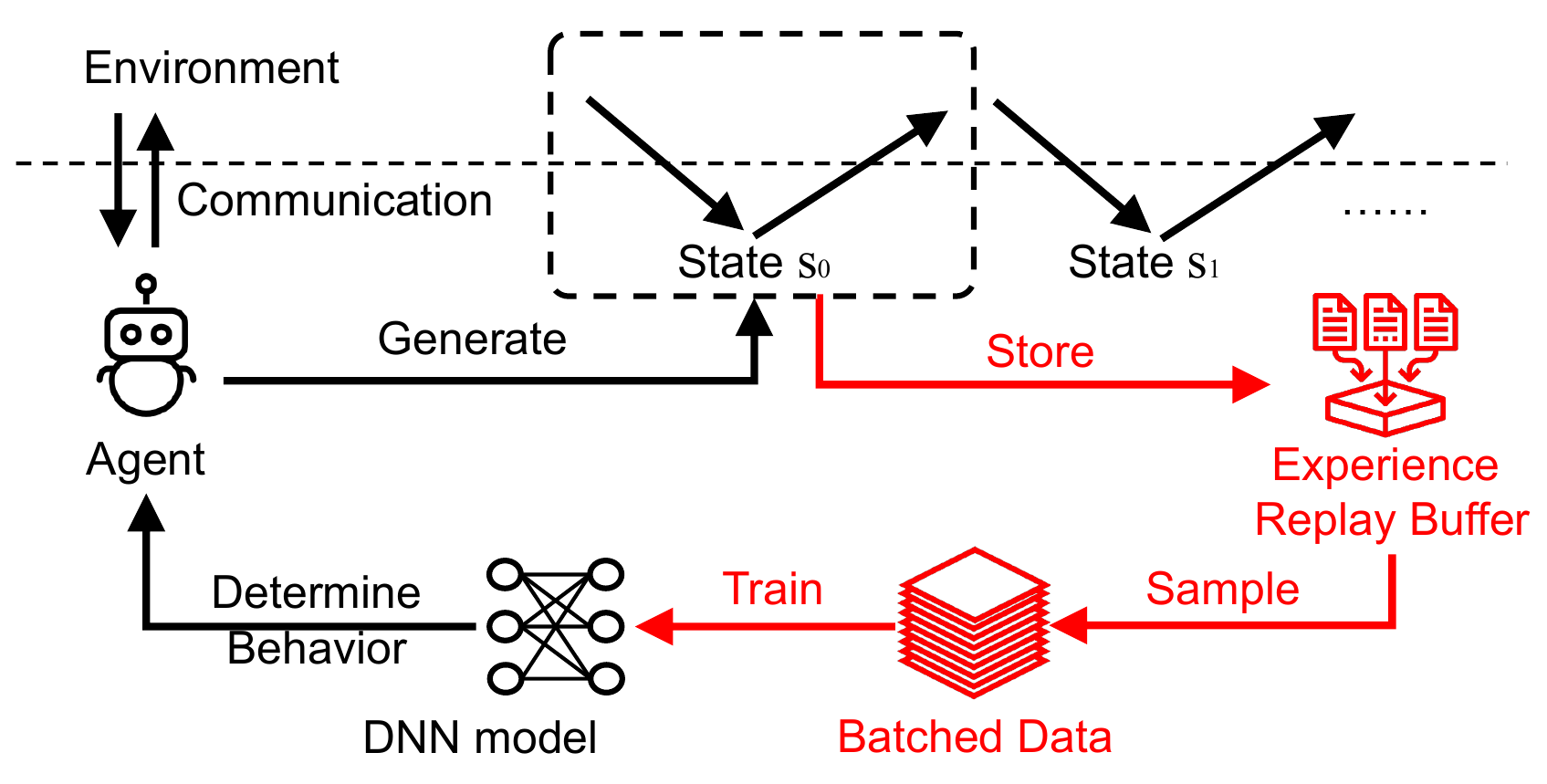}
\caption{An overview of Deep Reinforcement Learning (DRL).}
\label{fig:DRL_overview}
\vspace{-5mm}
\end{figure}

Deep Reinforcement Learning (DRL) is characterized by a combination of reinforcement learning (RL) algorithms and deep learning techniques  (as shown in Figure~\ref{fig:DRL_overview}), providing remarkable algorithm performance for extensive application prospects, particularly for intelligent robots~\cite{arulkumaran2017deep,chen2018tvm}.
However, integrating DRL solutions into real-world autonomous embedded system-driven robots would be more challenging because robots are required to continuously, efficiently, and frequently train and retrain DRL models in order to adapt to new environments with strict timing and resource limitations.
In such situations, meeting the real-time training deadlines is not merely a matter of budget adherence but a fundamental requirement to ensure system functionality, maintain up-to-date knowledge, and enhance accuracy. By emphasizing this, we make it clear that the timing sensitivity of DRL training extends beyond the offline workloads and is essential for a range of scenarios, especially those requiring real-time responses and continual learning. 

To illustrate this need, consider the following real-world examples: (1) Autonomous navigation robots: DRL models must constantly adapt to dynamic road environments in the context of autonomous navigation robots. They are often required to learn and retrain based on new data and evolving environments. In such cases, meeting deadlines during the training phase ensures that the autonomous navigation robots' decision-making system remains responsive and capable of handling new situations. (2) Search and Rescue Robots: In emergencies like disaster relief, robots need to navigate unpredictable challenges like damaged buildings or shifting debris. Quick training and updates to their DRL models allow them to adapt on-the-fly, ensuring they can effectively locate and assist those in need. (3) Drone Surveillance: Drones face a variety of challenges, from changing terrains in a forest to fluctuating weather conditions. Real-time DRL training ensures drones navigate these challenges efficiently, avoiding mishaps and ensuring effective monitoring.

To thoroughly comprehend the implications of Deep Reinforcement Learning (DRL) in embedded systems, it is essential to focus on two pivotal components (red parts in Figure~\ref{fig:DRL_overview}) of DRL training: batch execution and replay buffer~\cite{lillicrap2015continuous,mnih2015human,van2016deep,haarnoja2018soft,schulman2017proximal}. Batch execution involves processing multiple data samples in a single operation, fostering effective learning. The replay buffer, on the other hand, serves as a repository for past experiences, which are sampled for conducting minibatch training \cite{mnih2015human, mccloskey1989catastrophic}.
These components significantly impact not only the algorithm's application-level performance but also the system-level memory usage and latency \cite{horgan2018distributed,schaul2015prioritized,henderson2018deep}. This observation becomes particularly relevant in the context of real-time DRL systems, wherein training workloads consist of multiple time-bound tasks and subtasks, each encapsulating a complete DRL training episode. These tasks can be further divided into several subtasks, i.e., single-step-level minibatch training, with the flexibility for individual scheduling to meet specific end-to-end deadlines.

% System designers must carefully consider the trade-offs between parallelism, memory usage, and convergence time when selecting these characteristics for DRL training. The following sections will illustrate and explore this multi-dimensional tradeoff space for improving performance in resource-constrained scenarios.

\subsection{Balancing Data Parallelism and Memory Usage: A Focus on Training Batch Size}

\label{sec:case_study_batchsize}

%\Cong{Maybe it is clearer to directly correlate mem usage and latency, and explain data parallelism is the dominating factor behind how batching size impacts latency.}

In this subsection, we investigate the trade-off space between parallelism and memory usage in varying DRL training batch sizes, concentrating on its implications for both system-level and application-level performance. 
%\Cong{It is not clear what is your exp setup for conducting these case studies and why these studies reflect robotic scenarios.} 
All the data in case studies are based on  Classic Control~\cite{6313077}, which is a set of classical control games for DRL, capable of simulating realistic robots and complex mechanical systems. Classic Control problem emphasizes the importance of control theory concepts, which corresponds to the key control components in robotic scenarios contexts.

\begin{figure}[!t]
\centering
\begin{subfigure}[b]{0.240\textwidth}\includegraphics[width=\textwidth]{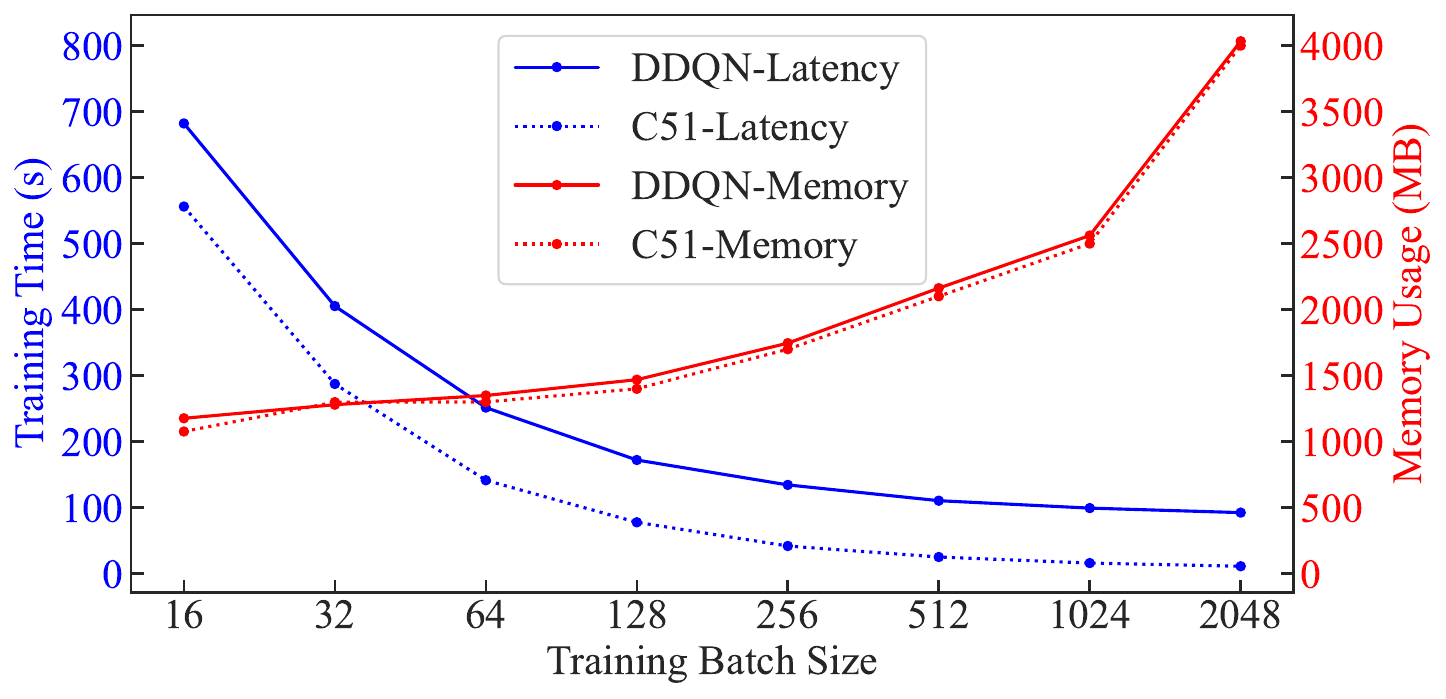}
\caption{Tradeoff for batch size}
\label{fig:batchsize_tradeoff}
\end{subfigure}
\begin{subfigure}[b]{0.225\textwidth}\includegraphics[width=\textwidth]{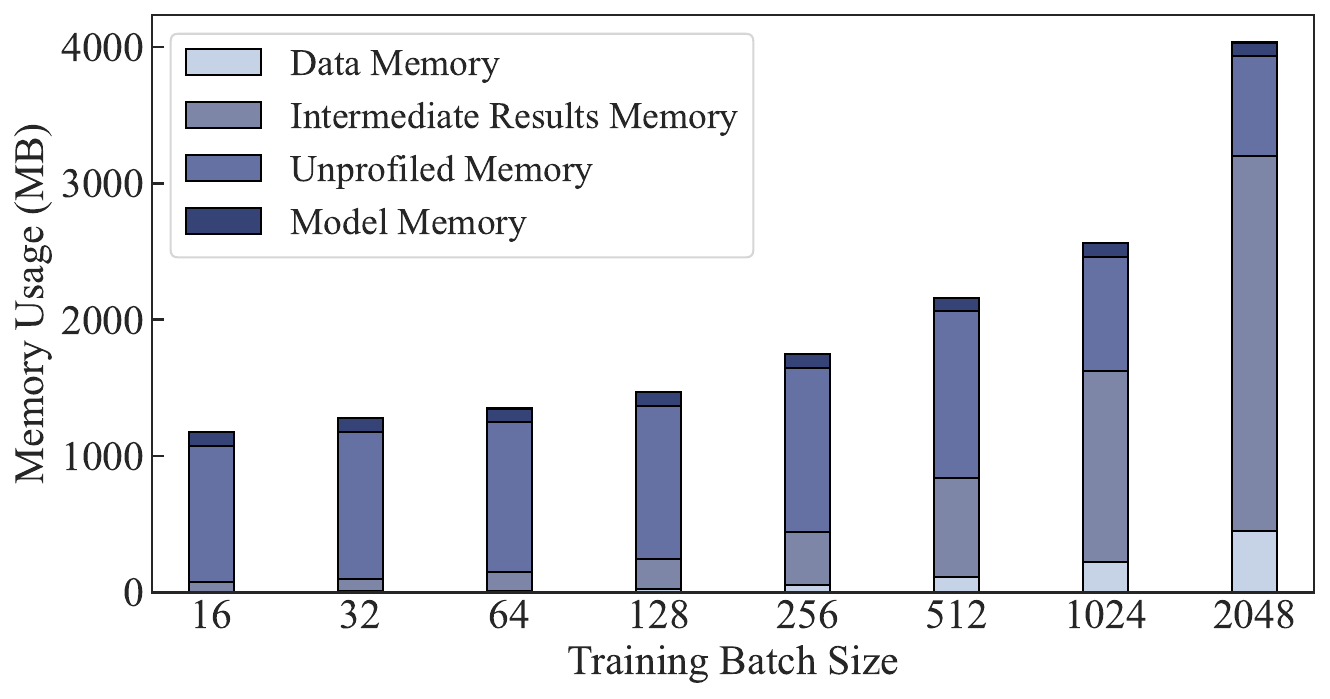}
\caption{Memory breakdown}
\label{fig:memory_breakdown}
\end{subfigure}
\caption{Balancing data parallelism and memory usage.}
\vspace{-3mm}
% \begin{subfigure}[b]{0.24\textwidth}\includegraphics[width=\textwidth]{figures/batchsize_memory.pdf}
% \caption{Memory Usage}
% \label{fig:batchsize_memory}
% \end{subfigure}
% \begin{subfigure}[b]{0.24\textwidth}\includegraphics[width=\textwidth]{figures/batchsize_throughput.pdf}
% \caption{Data Parallelism}
% \label{fig:batchsize_throughput}
% \end{subfigure}
% \begin{subfigure}[b]{0.3\textwidth}\includegraphics[width=\textwidth]{figures/batchsize_latency.pdf}
% \caption{Algorithm Performance}
% \label{fig:batchsize_latency}
% \end{subfigure}
% \caption{Analysis of memory usage in the autonomous learning library~\cite{nota2020autonomous} demonstrates that as training batch size increases, data and intermediate result memory requirements grow considerably, while model parameter memory remains minimal. Concurrently, with larger training batch sizes, the parallelism of DRL improves, reflected in a significant increase in data throughput per unit of time and a gradual decrease in convergence time. C51 refers to Categorical DQN defined in the autonomous learning library~\cite{nota2020autonomous}. \zexin{remove (c), don't mention algorithm performance in this case study}}
\label{fig:batchsize}
\end{figure}

As depicted in Figure~\ref{fig:batchsize}, we perform a comprehensive analysis of performance metrics based on the autonomous learning library~\cite{nota2020autonomous}, which offers insights into the impact of batch size on performance. 
Figure~\ref{fig:batchsize_tradeoff} highlights the trade-off space for training batch size, where data parallelism, the dominating factor affecting latency, can be improved with a larger training batch size, as long as it does not exceed the device's computing capability. However, as the training batch size increases, memory usage consistently grows for different replay-based DRL algorithms. In Figure~\ref{fig:memory_breakdown}, we provide a memory breakdown analysis, which shows that data and intermediate result memory requirements increase significantly with larger batch sizes while model parameter memory remains minimal.

Consequently, the optimal batch size depends on the specific algorithm and the available computing resources. A thorough examination and optimization of batch size are necessary for more efficient and effective DRL training in resource-constrained situations.

\noindent \textbf{Observation 1:} The trade-off between parallelism and memory usage is vital for optimizing DRL training. Finding the proper balance through batch size adjustments can enhance algorithm performance while minimizing computing resource consumption.

\begin{figure*}[!htbp]
    \centering
    \includegraphics[width=0.305\textwidth]{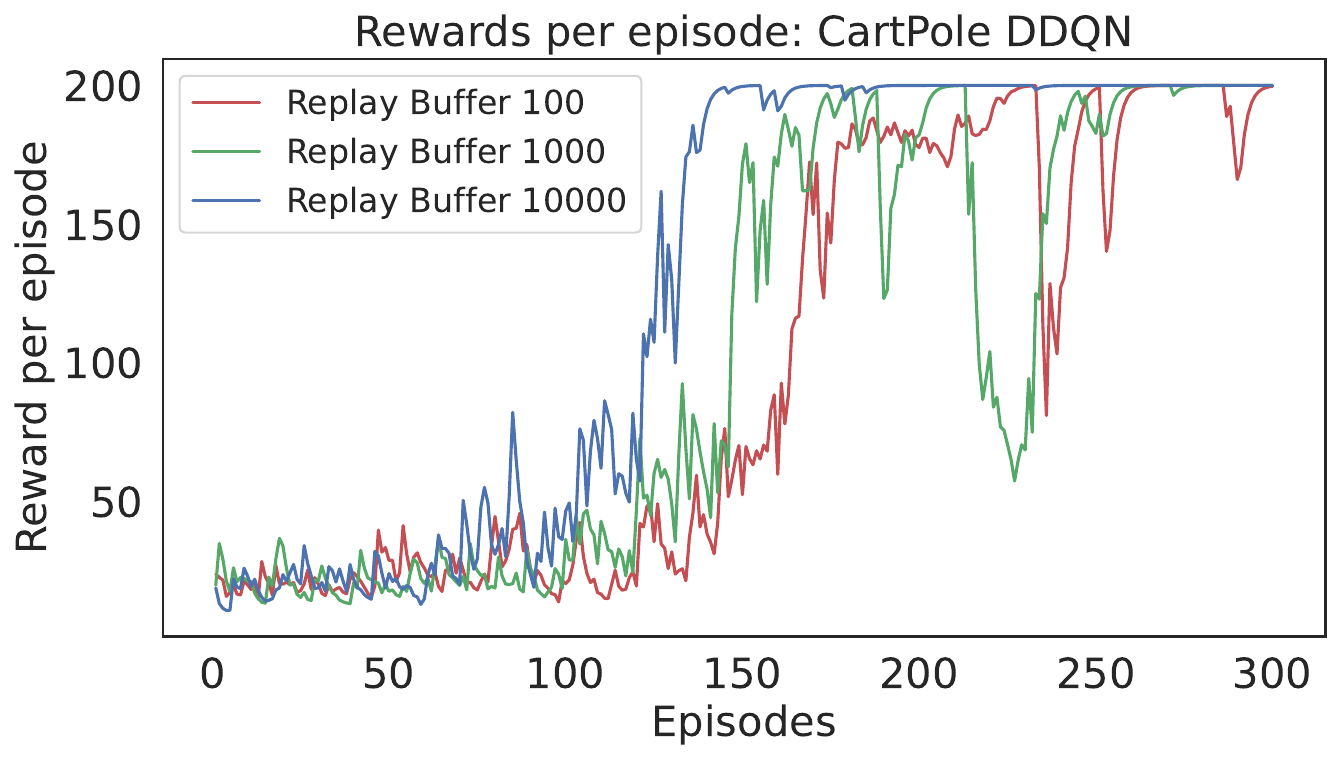}
    \includegraphics[width=0.305\textwidth]{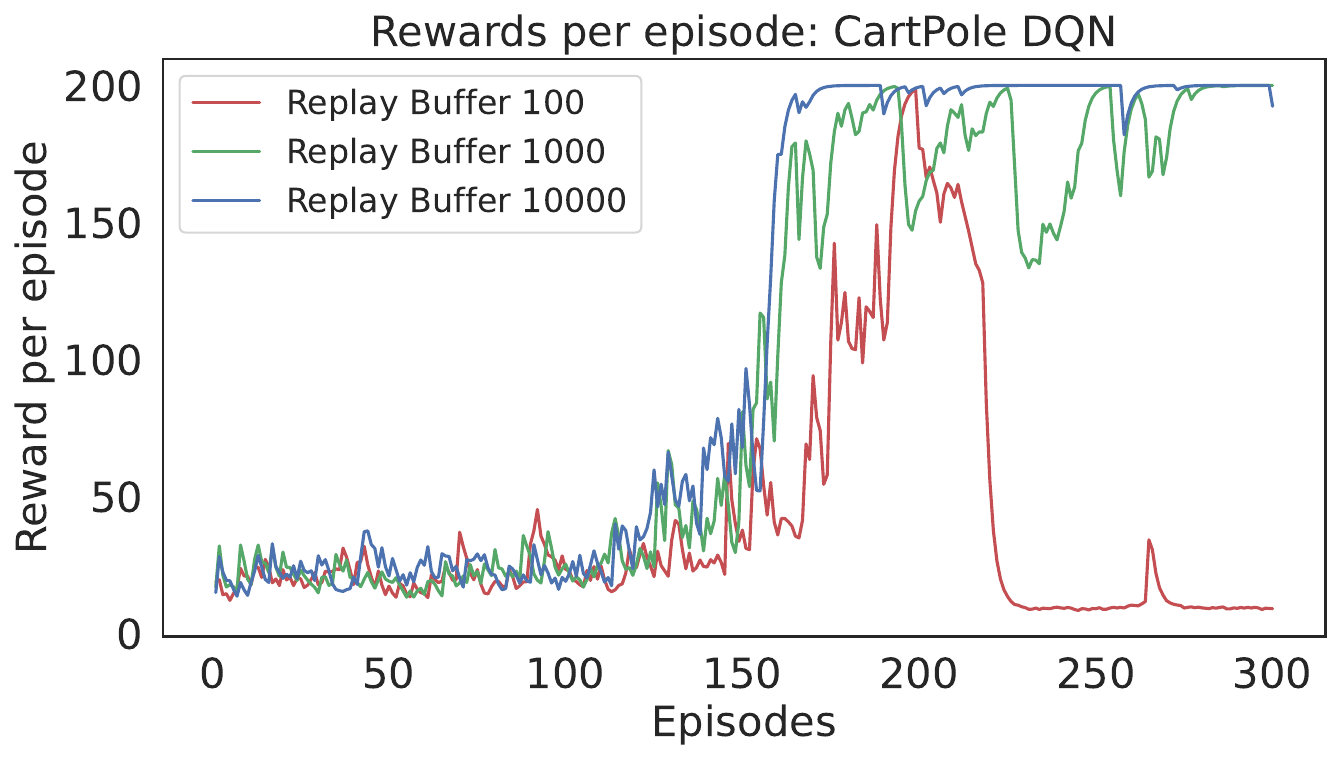}
    \includegraphics[width=0.305\textwidth]{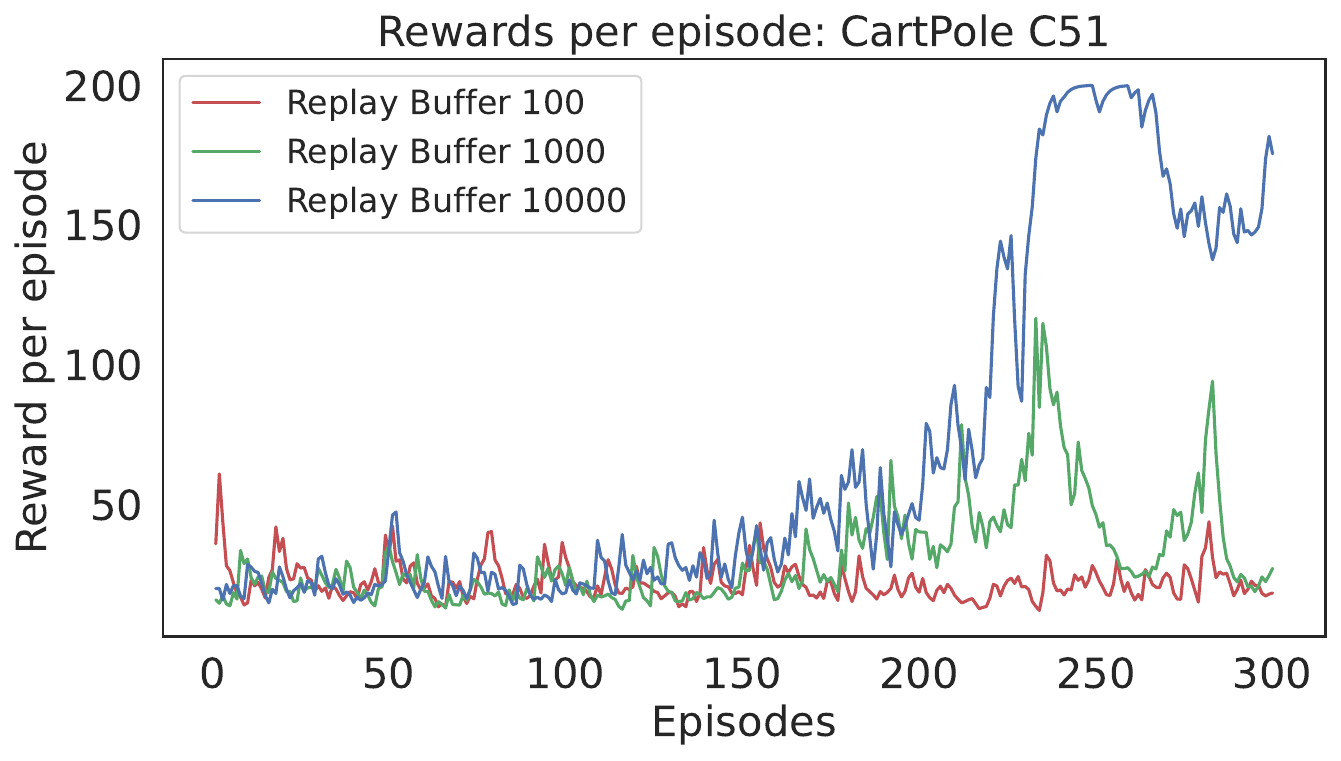}
    \caption{
    The deep reinforcement learning algorithm's performance (rewards) gradually increases as the replay buffer size increases and convergence of the algorithms becomes faster while the memory usage increases significantly across different algorithms. The experiments are running based on Classic Control~\cite{6313077} CartPole environment. C51 refers to Categorical DQN.}
    \label{fig:buffer_size}
    \vspace{-5mm}
\end{figure*}

\begin{table}[!t]
  \centering
  \caption{Common resource-oblivious DRL training settings may cause serious out-of-memory (OOM) on resource-constrained embedded devices. ``\checkmark'' refers OOM to happen.}
  \label{tab:models}

\renewcommand\arraystretch{1.3}
  \resizebox{0.45\textwidth}{!}{ 
    \begin{tabular}{|c|c|c|c|c|c|c|}
    \hline 
    \multirow{2}{*}{\textbf{Environment}} & \multirow{2}{*}{\textbf{Algorithm}} & \multirow{2}{*}{\textbf{Buffer Size}} & \multirow{2}{*}{\textbf{Batch Size}} & \multirow{2}{*}{\textbf{Peak Memory}}  & \multicolumn{2}{c|}{\textbf{Out-of-Memory}} \\
    \cline{6-7}
    & & & & & \textbf{Xavier}   & \textbf{Orin} \\
    \hline
    \multirow{3}{*}{Classic Control~\cite{6313077}} & DQN~\cite{mnih2015human} & 1,000,000 & 64 & 404.3MB & {\ding{55}} & {\ding{55}}\\
     & DDQN~\cite{van2016deep} & 1,000,000 & 64 & 404.3MB & {\ding{55}} & {\ding{55}}\\
     & C51~\cite{bellemare2017distributional} & 1,000,000 & 32 & 404.2MB & {\ding{55}} & {\ding{55}}\\
    \hline
    \multirow{3}{*}{Atari~\cite{bellemare2013arcade}} & DQN~\cite{mnih2015human} & 1,000,000 & 32 & 40384.5MB & {\checkmark} & {\checkmark}\\
     & DDQN~\cite{van2016deep} & 1,000,000 & 32 & 40384.5MB & {\checkmark} & {\checkmark}\\
     & C51~\cite{bellemare2017distributional} & 1,000,000 & 32 & 40384.5MB & {\checkmark} & {\checkmark}\\
    \hline
    \multirow{3}{*}{DonkeyCar~\cite{bib:gymdonkeycar}} & DQN~\cite{mnih2015human} & 100,000 & 128 & 35522.6MB & {\checkmark} & {\checkmark} \\
    & DDQN~\cite{van2016deep} & 100,000 & 128 & 35122.6MB & {\checkmark} & {\checkmark} \\
    & C51~\cite{bellemare2017distributional} & 100,000 & 64 & 33329.8MB & {\checkmark} & {\checkmark}\\
    \hline
    \end{tabular}
}
  \label{tab:oom}
  \vspace{-5mm}
\end{table}

% \begin{figure*}[!htbp]
%     \centering
%     \begin{subfigure}[b]{0.60\textwidth}
%     \includegraphics[width=0.32\textwidth]{figures/conflict_algorithm_performance_new.pdf}
%     \includegraphics[width=0.32\textwidth]{figures/conflict_data_parallelism_new.pdf}
%     \includegraphics[width=0.32\textwidth]{figures/conflict_memory_efficiency_new.pdf}
%     \caption{A three-dimensional analysis of performance metrics in embedded deep reinforcement learning to visualize the contribution of influence.}
%     \label{fig:no_silver_bullet}
%     \end{subfigure}
%     \hspace{10px}
%     \begin{subfigure}[b]{0.3\textwidth}
%     \includegraphics[width=\textwidth]{figures/simple_opt_conflit.pdf}
%     \caption{Simply combination fails in memory-limited autonomous embedded scenarios. 
%     }
%     \label{fig:simple_opt_conflit}
%     \end{subfigure}
    
%     \caption{\textbf{No silver bullet.} (a) Navigating the trade-offs: a three-dimensional analysis of performance metrics in embedded deep reinforcement learning and the challenge of out-of-memory issues. All data are normalized to the interval from zero to one to visualize the contribution of influence better. Best viewed in color. (b) An intuitive example to show the conflict when simply combining the two 2-dimensional spaces on memory-limited autonomous embedded systems (AES). \color{red}{consider delete figure (b) by rw \#4}}
%     \vspace{-5mm}
% \end{figure*}

\begin{figure}[!htbp]
    \centering
    \includegraphics[width=0.155\textwidth]{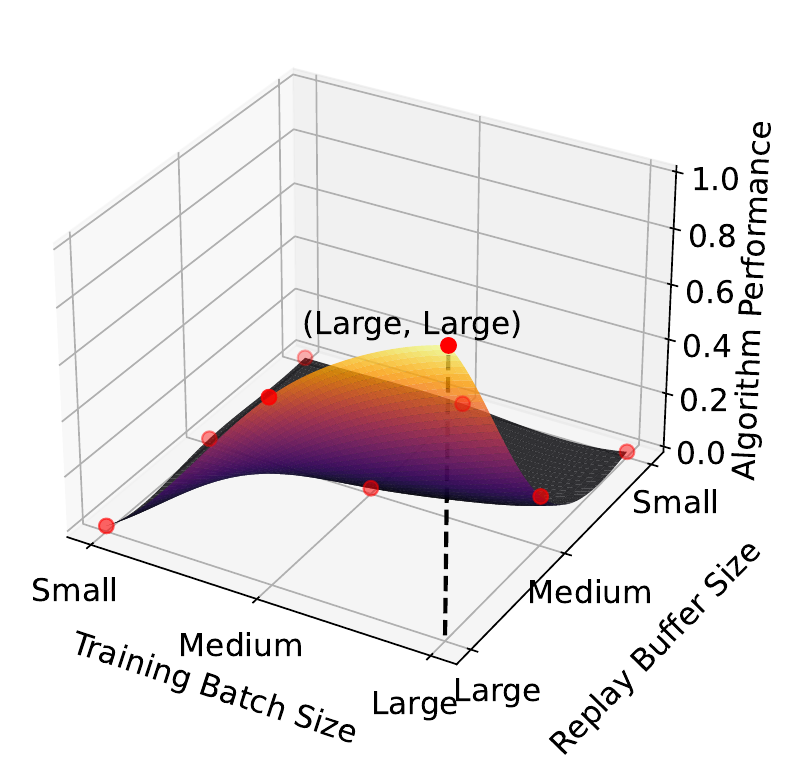}
    \includegraphics[width=0.155\textwidth]{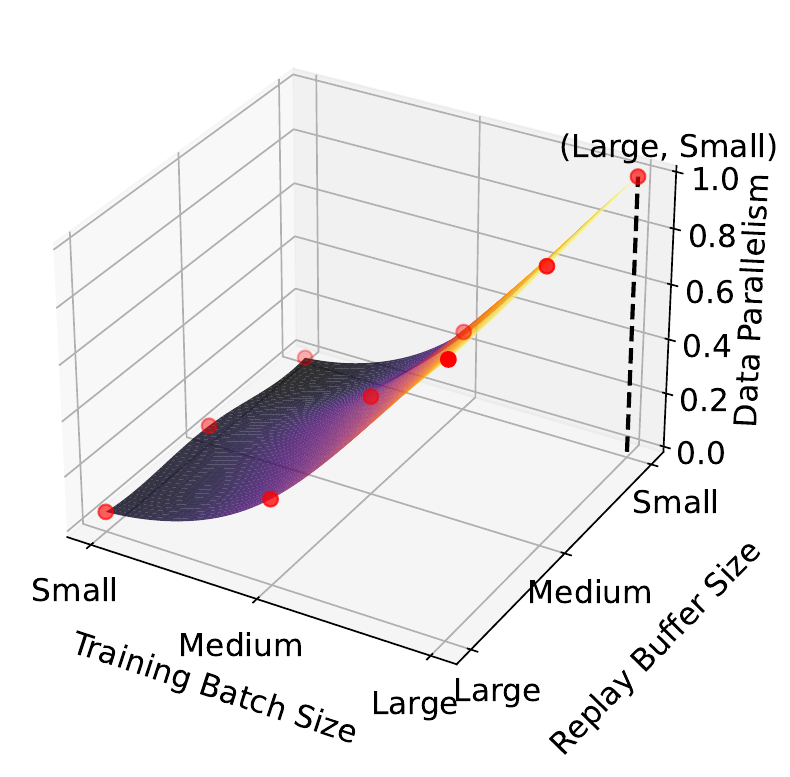}
    \includegraphics[width=0.155\textwidth]{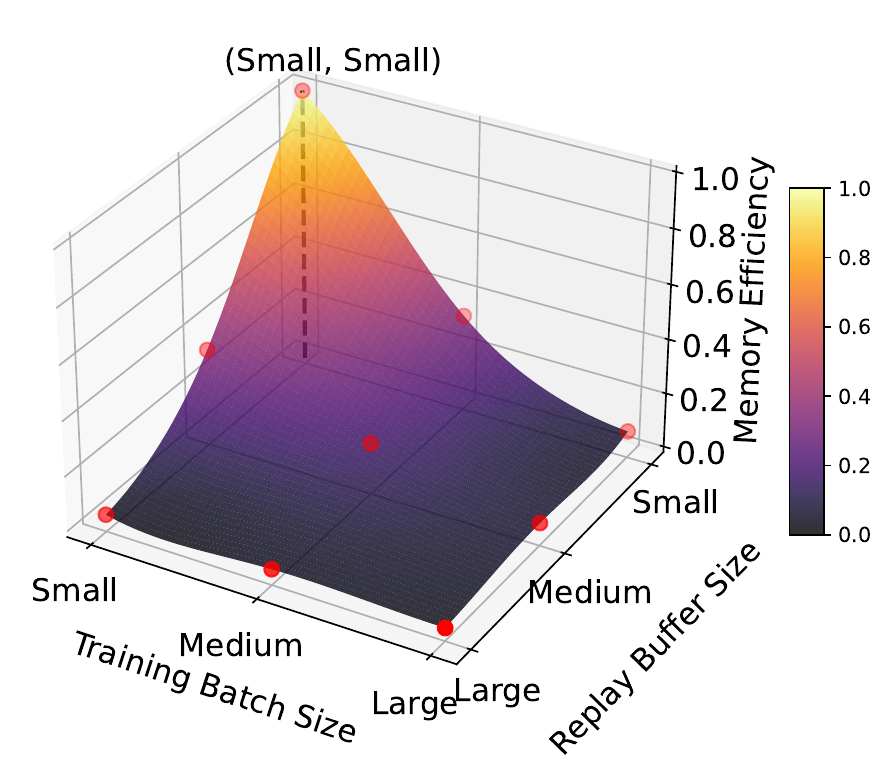}
    \caption{\textbf{No silver bullet.} Navigating the trade-offs: a three-dimensional analysis of performance metrics in embedded deep reinforcement learning and the challenge of out-of-memory issues. All data are normalized to the interval from zero to one to visualize the contribution of influence better.}
    \label{fig:no_silver_bullet}
    \vspace{-5mm}
\end{figure}

\subsection{Balancing Algorithm Performance and Memory Usage: A Focus on Replay Buffer Size}

\label{sec:case_study_replaybuffer}

In this subsection, we explore the trade-offs between algorithm performance and memory usage in DRL training with respect to replay buffer size. Figure~\ref{fig:buffer_size} illustrates the clear tradeoff space between algorithm performance, exemplified by convergence time, and memory usage in DRL scenarios utilizing replay buffers. As the replay buffer size increases, the algorithm performance generally improves, leading many DRL training approaches to adopt large buffer sizes to boost algorithm performance. Additionally, cumulative rewards gained from DRL training rise with larger replay buffer sizes.

Nonetheless, it is essential to acknowledge that expanding replay buffer sizes also entails increased memory usage, potentially causing out-of-memory (OOM) issues, particularly in memory-constrained embedded devices. Table~\ref{tab:oom} presents default settings for various DRL benchmarks and examines the peak memory allocation for the buffer. Notably, some default settings for widely-used DRL algorithms could trigger OOM in powerful NVIDIA embedded devices.

\noindent \textbf{Observation 2:} Balancing algorithm performance and memory usage for replay buffer size are crucial in DRL training, particularly for memory-constrained embedded devices. A thorough examination of characteristics and trade-offs enables the optimization of performance while meeting practical memory constraints.

\subsection{Managing Complex Trade-offs under Practical System Scenarios: Balancing in the Three-dimensional Space}

\label{sec:case_study_3d}

%\Cong{Best to add an clean, intuitive example showing the confliction when simply combining the two 2-dimensional spaces.}

In light of the complexities of DRL and the trade-offs explored in the previous subsections, we now focus on the broader challenge of balancing conflicting characteristics in the three-dimensional space of DRL training under practical system scenarios.

As depicted in Figure~\ref{fig:no_silver_bullet}, interestingly, patterns about the batch size and replay buffer size in different performance dimensions display vastly different behavior. This stark difference implies that DRL inherently involves conflicting characteristics. Simply co-optimizing some of these aspects may result in suboptimal solutions, underscoring the need for a more comprehensive and integrated approach. Moreover, practical challenges arise when considering resource-constrained embedded devices, where memory constraints may become a significant issue. Recall that simply combining the two 2-dimensional space optimization (as discussed in Sec.\ref{sec:case_study_batchsize} and Sec.\ref{sec:case_study_replaybuffer}) could exacerbate out-of-memory (OOM) problems.

 % As shown in Figure~\ref{fig:R3_challenge_overview}

To address such challenges, it is essential to carefully consider all characteristics in the three-dimensional space and their trade-offs while also taking into account the practical constraints of embedded devices. By adopting such a holistic approach, DRL training can achieve near optimal performance while managing these conflicting characteristics under practical system scenarios.

\noindent \textbf{Observation 3:} Managing trade-offs between various DRL training characteristics is a complex task due to multiple conflicting factors. A comprehensive approach considering all characteristics and performance objectvies is necessary, particularly for resource-constrained embedded devices. 

\section{Methodology}

\subsection{System Overview}

\begin{figure}[!htbp]
\centering
\includegraphics[width=0.5\textwidth]{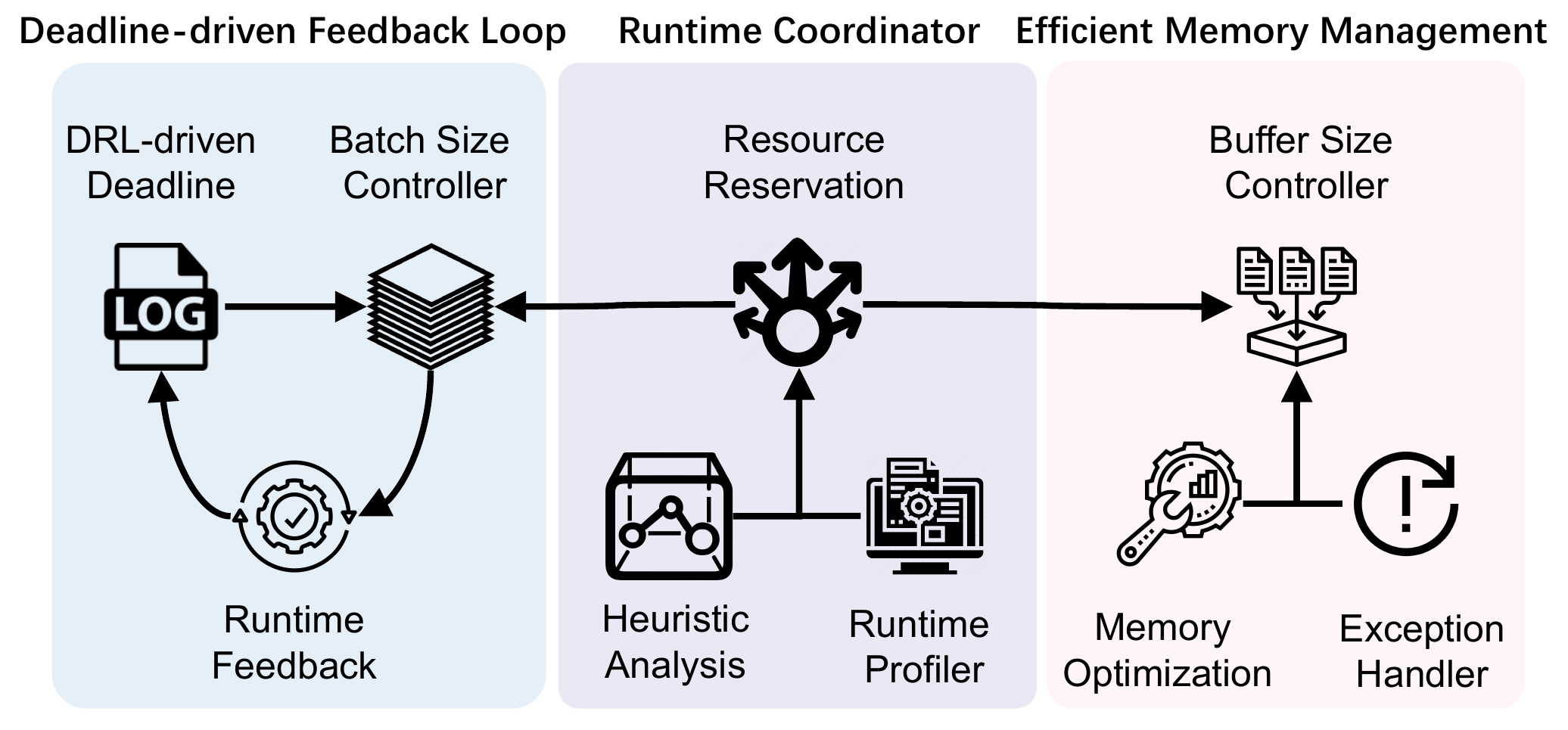}
\caption{An overview of \Approach{}. }
\label{fig:overview}
\vspace{-5mm}
\end{figure}

In response to the unique challenges of \textbf{R}eal-time deep \textbf{R}einforcement learning for \textbf{R}obotics outlined in Sec.~\ref{sec:case_study}, we propose a comprehensive framework \Approach{}, illustrated in Figure~\ref{fig:overview}. This framework encompasses two main components: a deadline-driven feedback loop and efficient memory management, which together facilitate the balance between the two-dimensional trade-off of batch size and replay buffer size. Moreover, a runtime coordinator is introduced to manage the complex three-dimensional trade-off under resource-constrained scenarios, ensuring the adaptation of different performance goals without conflicts. Each component is discussed in further detail below:

\begin{itemize}[leftmargin=10px]
\item \textbf{Deadline-driven Feedback Loop:}
This component aims to balance memory constraints and latency (Sec.~\ref{sec:case_study_batchsize}). It dynamically assigns an intermediate deadline for each training episode based on the computation dynamics nature of DRL (as illustrated in Figure~\ref{fig:computational_dynamic}). Subsequently, it determines the optimal batch size that adheres to memory constraints while meeting the subtask deadline for the current training episode. If every subtask completes by its assigned intermediate deadline, the end-to-end deadline will be met.

\item \textbf{Efficient Memory Management:}
In addressing the challenges posed by memory-constrained embedded systems for DRL, efficient memory management is vital. Our proposed solution, corresponding to the trade-off discussed in Sec.~\ref{sec:case_study_replaybuffer}, implements task-specific memory optimizations to significantly reduce memory footprint. This enables larger replay buffer sizes for enhanced algorithm performance in real-time embedded DRL. Furthermore, we design an exception handler to eliminate unexpected out-of-memory (OOM) occurrences. Based on these underlying subcomponents, the buffer size controller can decide on demand how to assign replay buffer size.

\item \textbf{Runtime Coordinator:}
The runtime coordinator is essential for the seamless operation of \Approach{}, ensuring real-time DRL by synchronizing the interactions between the deadline-driven feedback loop, memory management, and other system components. Utilizing analytical modeling and a runtime profiler, the coordinator dynamically adjusts the resource reservation strategy in accordance with the actual system resource constraints. This guarantees that the runtime coordinator expertly maintains the delicate balance of factors required for achieving optimal DRL agent training.
\end{itemize}

By coherently integrating these three components, \Approach{} provides a high-performance solution for training deep reinforcement learning agents in real-time. This enables rapid adaptation to dynamic and complex environments. 

\subsection{Deadline-driven Feedback Loop}

\label{sec:deadline_driven_feedback}

As emphasized in the preceding case study (Sec.~\ref{sec:case_study_batchsize}), achieving a balance between data parallelism and memory usage is vital for optimizing embedded DRL. 
Our goal is to meet end-to-end deadlines within hard memory constraints by adjusting the batch size. Intuitively, one possible approach is to partition subtasks and assign an appropriate intermediate deadline to each subtask. However, before designing a corresponding solution, it is necessary to consider how to partition the DRL workload and assign the intermediate deadline.

\begin{figure}[!htbp]
\centering
\begin{subfigure}[b]{0.240\textwidth}\includegraphics[width=\textwidth]{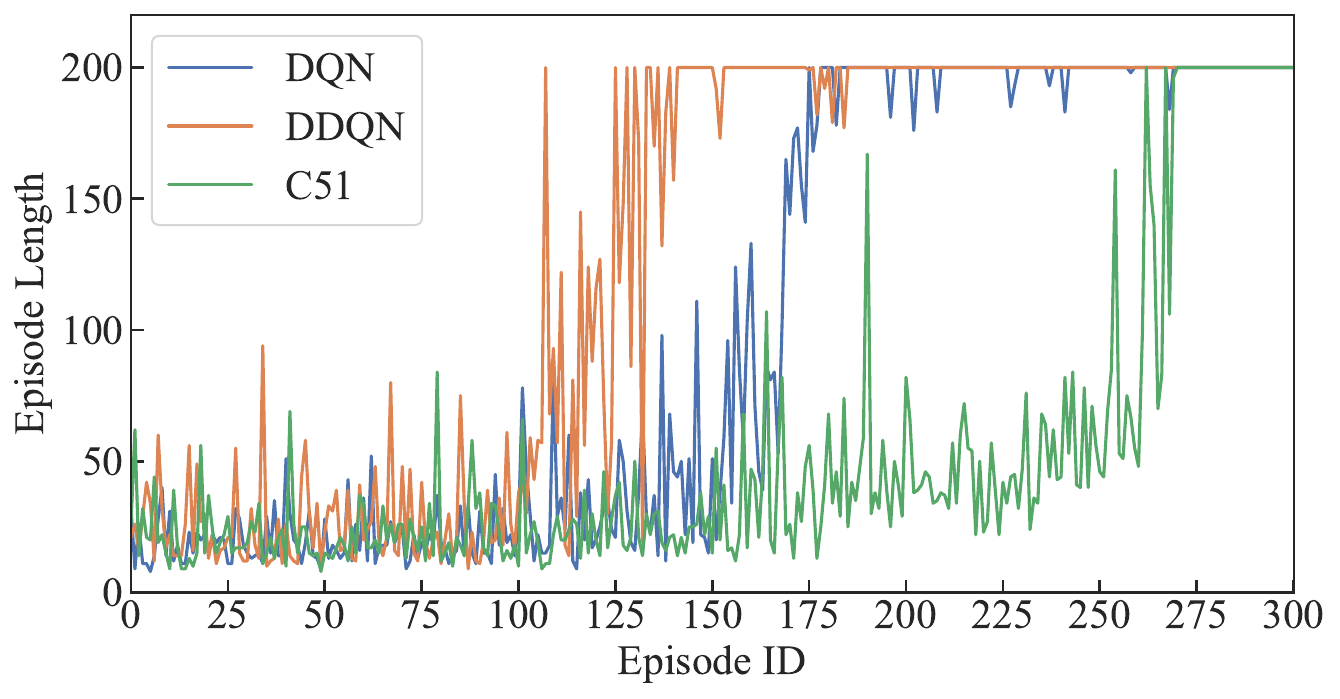}
\caption{Workload-induced dynamics}
\label{fig:workload_dynamics}
\end{subfigure}
\centering
\begin{subfigure}[b]{0.230\textwidth}\includegraphics[width=\textwidth]{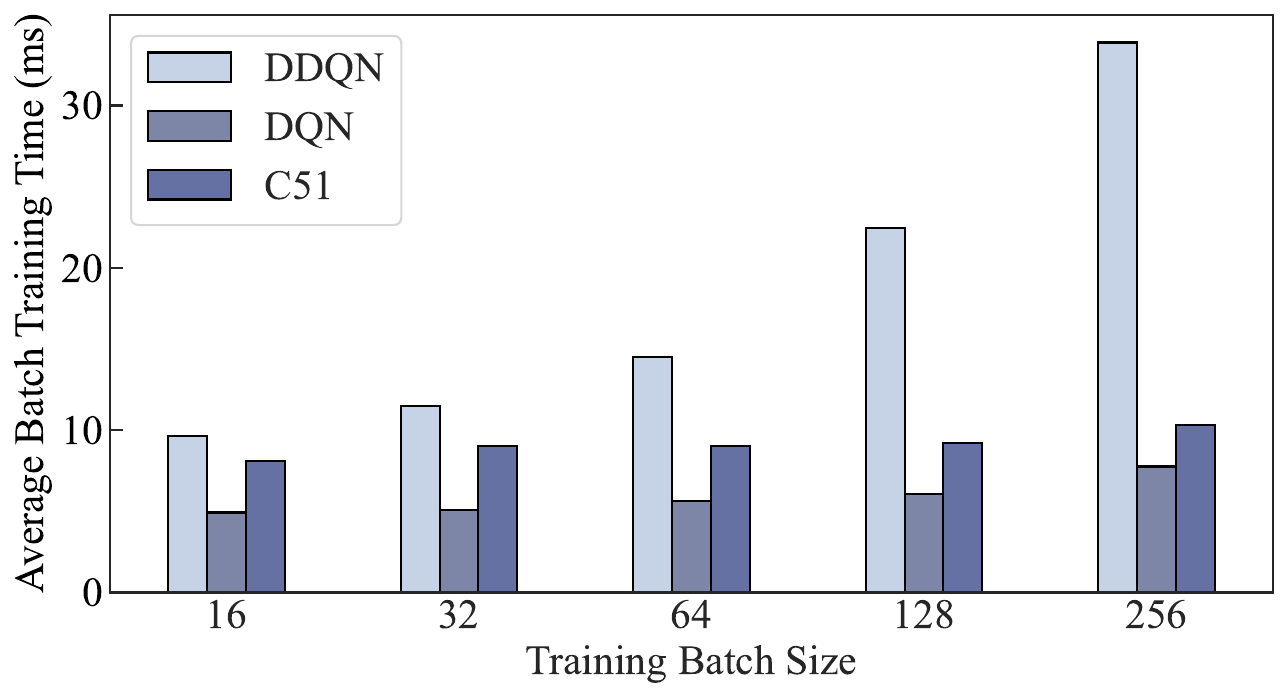}
\caption{Batching-induced dynamics}
\label{fig:batching_dynamics}
\end{subfigure}
\caption{The computational dynamic nature of DRL emphasizes the need for dynamic intermediate deadline assignments.}
\vspace{-3mm}
\label{fig:computational_dynamic}
\end{figure}

Although DRL can be naturally divided into episodes, the computational dynamics of DRL workload add complexity to the assignment of reasonable intermediate deadlines. As illustrated in Figure~\ref{fig:computational_dynamic}, we leverage an example of DRL classic control~\cite{6313077} to show this point in two folds.
At the DRL workload level, the computational cost of each natural computational subdivision unit (episode) varies over time, as depicted in Figure~\ref{fig:workload_dynamics}. The episode length was short at the beginning of DRL training; as the training progressed, the episode length gradually increased. Furthermore, examining finer-grained computation reveals that, since we aim to dynamically adjust the batch size, the computational cost of each batch (step) is also distinct, as depicted in Figure~\ref{fig:batching_dynamics}. This computational nature presents a further challenge to intermediate deadline assignments, which is hard to model without timing-consuming profiling. This implies that straightforwardly conducting a modeling-based optimization for all three dimensions may be infeasible across different DRL algorithms. 

To address such a challenge, we propose an intuitive, practical deadline driven feedback loop mechanism that adapts to the varying computational dynamics of DRL workloads. The intuition is to directly adjust the batch size based on progress tracking without explicitly assigning intermediate deadlines. Specifically, we design to periodically assess the remaining time to dynamically adjust the training batch size incorporating temporal feedback information and thus meet the end-to-end deadline. Formally, as the DRL training process unfolds, two critical constraints often imposed are a data budget ($B$) and an end-to-end deadline ($D$). We keep track of the elapsed training time ($t_i$) and consumed budget ($s_i$) at the start of each episode $i$. We employ two trackers to monitor the progress of the DRL process: a timing tracker evaluates the ratio of elapsed training time to the given deadline, and a data tracker monitors the proportion of the consumed budget to the total budget. These ratios, denoted as $a$ and $b$ respectively, are given by $a = t_i / D$ and $b = s_i / B$.

These ratios serve as comparative indicators to guide the adjustment of the DRL process. If $a > b$, it indicates that the DRL training is lagging behind, necessitating acceleration by multiplying the current batch size by a scale factor $c$. Conversely, if $a < b$, it suggests the training is on track, thereby allowing for deceleration by dividing the batch size by the scale factor $c$. However, to maintain the stability of the DRL process, it is crucial to prevent the batch size from becoming excessively small or large.~\cite{mnih2015human,schulman2017proximal,lillicrap2015continuous} As such, we confine the batch size within a specified range as per established DRL practices as follows:

\begin{equation}
b_{i+1} = \min( \max(b_i, b_{\min}), 4 * b_{\min}).
\label{eq:stable}
\end{equation}

Ultimately, it is critical to ensure that the DRL process complies with the hard memory constraint given by:% \aritra{Maybe you can change the equation to say b(i+1) = min(b(i), ...)}:

\begin{equation}
b_{i+1} = \min\left(b_{i+1}, M_{batch} \cdot \frac{b_{base}}{M_{base}}\right).
\label{eq:linear_scaling}
\end{equation}

Two key components in this context are $M_{base}$ and $M_{batch}$, both playing essential roles in memory reservation for batch execution during the scaling of minibatch size.
The $M_{base}$ represents the memory reserved for batch execution considering a base minibatch size, denoted as $b_{base}$. This reservation is determined by the initial episode hyperparameter configuration, thereby serving as a reference point for subsequent adjustments.
Conversely, $M_{batch}$ signifies the memory currently reserved for batch execution, a value that can dynamically vary in accordance with system constraints and requirements.

To further enhance the responsiveness, we integrate granular step-level adjustments within our feedback loop, enabling meticulous progress tracking at each training step for precise monitoring of elapsed time and budget expenditure. The detailed structure of this fine-grained deadline feedback loop aligns with the episode-level setup, facilitating seamless implementation without extensive modifications. Although this granular control promotes rapid adjustment to computational dynamics, it potentially introduces additional overhead that may affect training performance. Hence, the selection between episode-level and step-level adjustments is contingent upon specific DRL task requirements and available resources.

\subsection{Efficient Memory Management}

\label{sec:effcient_memory_managemnet}

As demonstrated in Sec.~\ref{sec:case_study_replaybuffer}, a larger replay buffer enhances algorithm performance but also demands more memory. The performance of DRL algorithms relies heavily on the quantity of data, which in turn depends on available memory resources. This is particularly relevant in resource-constrained embedded systems, where memory resources are limited and must be meticulously managed to avoid out-of-memory (OOM) issues. To address this challenge, we propose an efficient memory management scheme that contains memory optimization, an exception handler, and an application-level buffer size controller.

\begin{figure}[!htbp]
\centering
\begin{subfigure}[b]{0.240\textwidth}\includegraphics[width=\textwidth]{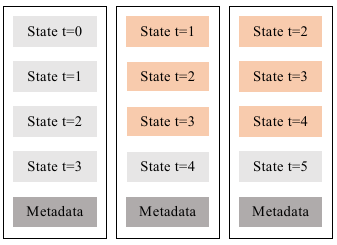}
\caption{Dummy replay buffer}
\label{fig:dummy_replay_buffer}
\end{subfigure}
\centering
\begin{subfigure}[b]{0.240\textwidth}\includegraphics[width=\textwidth]{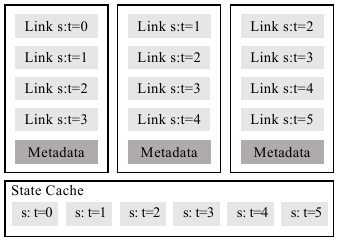}
\caption{Efficient replay buffer}
\label{fig:efficient_replay_buffer}
\end{subfigure}
\caption{Replay buffer memory optimization. The orange color indicates redundant data in the dummy replay buffer.}
\vspace{-3mm}
\label{fig:replay_buffer}
\end{figure}

First, to target real-time embedded systems with stringent memory constraints, we propose underlying system-level DRL-task-specific optimizations for efficient memory management. First, we conduct a finer-grained analysis of the replay buffer and observe that the current caching mechanism of DRL training is inefficient. As illustrated in Figure~\ref{fig:dummy_replay_buffer}, each data frame is stored in a tuple data structure (states, metadata) in the replay buffer, recording multiple time steps. It has to be admitted that such a dummy implementation can avoid frequent memory accesses and thus achieve higher parallelism, however, it would result in significant memory redundancy due to duplicated data shown in orange color. Ideally, each time step should maintain only one copy of data in the replay buffer. Based on this observation, we design and implement a simple yet efficient caching method that significantly reduces the memory footprint of the replay buffer, allowing for larger buffer sizes within memory-constrained embedded devices. As shown in Figure~\ref{fig:efficient_replay_buffer}, we use a practical example of the Atari Breakout benchmark~\cite{mnih2015human} to illustrate the ideal memory efficiency of our method. For each data frame of the dummy replay buffer implementation, each data frame stores 4 consecutive state information (i.e., 4 moments of RGB images with size (210,160), occupying size 4x210x160x3=403,200 bytes). In contrast, the metadata, containing action, reward, and done flag, occupies very little memory space (9 bytes) in total. Therefore, we use a contiguous memory for storing state information, while only storing soft links (4 bytes for each link) in the data frame. When the corresponding data frame is required, the corresponding tuple data is filled on demand. Our efficient replay buffer can ideally achieve about 75\% less memory assumption to support the same length as a dummy replay buffer as in Figure~\ref{fig:replay_buffer}, because in most data frames, three out of four pieces (orange) of state information are replaced with lightweight soft links.

Let $M_{replay}$ represent the total memory reserved for the replay buffer, $S_{state}$ represent the size of one state, $S_{metadata}$ represent the metadata size, and $N_{frames}$ represent the number of frames stored in a single data frame (in the Atari Breakout example, $N_{frames} = 4$). The replay buffer size $r_{i}$ for training episode $i$ can be calculated as Eq.~\ref{eq:max_replay_buffer_size}, where $S_{link}$ denotes the size of a soft link, which is used to reference the state information in the contiguous memory:

\begin{equation}
\begin{aligned}
r_i = max\{ M_{replay}, (N_{frames} + i - 1) * S_{state} + \\
(N_{frames} - 1) \cdot i \cdot S_{link} + i \cdot S_{metadata}\}
\end{aligned}
\label{eq:max_replay_buffer_size}
\end{equation}

Directly applying this efficient caching method could reduce memory redundancy, however, it could introduce substantial overhead since it necessitates more memory access, potentially harming data parallelism in DRL training. Inspired by recent work~\cite{DBLP:conf/rtss/JiYKASDK22}, we optimize the training process with non-blocking data prefetching to hide latency. These task-specific optimizations enable real-time embedded DRL training to achieve larger replay buffer sizes for improved algorithm performance while minimizing side effects on data parallelism.
To handle the possibility of unexpected out-of-memory (OOM) issues caused by memory interference or other unforeseen circumstances, we propose to design an exception handler mechanism. This mechanism can monitor memory usage during runtime and promptly react when an OOM exception is detected. The handler can temporarily scale down the memory reservations for both batch execution and replay buffer, which allows the system to recover gracefully from the OOM situation. Furthermore, it can record the OOM events and adjust the memory allocation strategy accordingly, ensuring the system remains robust under memory interference.

Specifically, we show part of the code example of a standard DDQN training step (as shown in Listing 1). We use a multi-threading technique to hide the latency of data prefetching (line 4) and thus achieve a better training step latency. Based on the above optimizations, the buffer size controller could calculate the current maximum replay buffer size based on the given memory limit and dynamically allocate or free memory at runtime.

\begin{lstlisting}[language=Python, caption=An example code of DRL training.]
def _train(self):
  if self._should_train():
    # sample transitions from buffer
    (states, actions, rewards, next_states, weights) = self.replay_buffer.sample(self.minibatch_size)
    # forward pass
    values = self.q(states, actions)
    # compute targets and loss
    next_actions = torch.argmax(self.q.no_grad(next_states), dim=1)
    targets = rewards + self.discount_factor * self.q.target(next_states, next_actions)
    loss = self.loss(values, targets, weights)
    # backward pass
    self.q.reinforce(loss)
    # update replay buffer
    self.replay_buffer.update()
\end{lstlisting}
\vspace{-3mm}

\subsection{Runtime Coordinator}

\label{sec:runtime_coord}

The Runtime Coordinator in \Approach{} is a pivotal component devised to address the intricate triadic trade-off between algorithm performance, memory efficiency, and data parallelism (as outlined in Sec.~\ref{sec:case_study_3d}). This component serves to orchestrate the execution of various system aspects, including the deadline-driven feedback loop and efficient memory management, thereby enabling smooth real-time DRL training. The coordination is executed through heuristic analysis, taking into consideration the outputs from the aforementioned system components alongside system-level resource constraints and performance objectives.

Memory reservations $M_{batch}$ and $M_{replay}$ are adjusted using a heuristic formulation that factors in the system's current performance and memory constraints. For episode $i$, the episode runtime is denoted as $t_i$, and the cumulative reward is $R_i$. Two scaling factors, $\alpha$ and $\beta$, are employed to fine-tune the memory reservations based on the system's performance:

\begin{equation}
\begin{aligned}
\alpha &= \frac{\gamma t_i}{\sum\limits_{j=1}^{\gamma}{t_{i - j}}} \\
\beta &= \frac{\gamma R_i}{\sum\limits_{j=1}^{\gamma}{R_{i - j}}}
\end{aligned}
\vspace{-2mm}
\end{equation}

Here, $t_i$ is the runtime of episode $i$, $R_i$ is the cumulative reward at the i-th episode, $\gamma$ is a hyperparameter, and $R_{i - \gamma}$ is the cumulative reward $\gamma$ steps prior to the i-th episode. $\gamma$'s role is to control the reward comparison span. By modulating $\gamma$, the algorithm can be adjusted to focus on short-term or long-term reward trends, influencing the training process and potentially enhancing overall performance. $\alpha$ and $\beta$ track latency and algorithm performance dynamically. To this end, the memory reservations for batch execution and replay buffer are then updated based on $\alpha$ and $\beta$ :
\vspace{-3mm}

\begin{equation}
\begin{aligned}
&M_{batch}^{new} = M_{batch} \cdot (1 + \max(\alpha-1, 0) \cdot (1 - \min(\beta,1))) \\
&M_{replay}^{new} = M_{replay} \cdot (1 + \min(\alpha,1) \cdot \max(1-\beta,0))
\end{aligned}
% \vspace{-2mm}
\end{equation}

Here, $M_{batch}^{new}$ and $M_{replay}^{new}$ represent the updated memory reservations for batch execution and replay buffer, respectively. This adjustment mechanism increases batch memory if the episode runtime is shorter than the subtask deadline and the cumulative reward is increasing relative to the training loss, while reducing replay buffer memory. To ensure total memory usage does not exceed system constraints, the sum of $M_{batch}^{new}$ and $M_{replay}^{new}$ must be checked against the available memory budget. If it exceeds the limit, memory reservations are proportionally scaled down to fit within the constraints.
Let's denote $M$ as the total available memory. We want to proportionally distribute $M$ between $M_{batch}$ and $M_{replay}$. First, we calculate the proportion of each reservation to the total desired memory, i.e., the sum of $M_{batch}^{new}$ and $M_{replay}^{new}$. Then, we allocate the memory in proportion to these ratios. The resulting equations are:

\begin{equation}
\begin{aligned}
    M_{batch}^{final} = M \cdot \frac{M_{batch}^{new}}{M_{batch}^{new} + M_{replay}^{new}} \\
M_{replay}^{final} = M \cdot \frac{M_{replay}^{new}}{M_{batch}^{new} + M_{replay}^{new}}
\end{aligned}
\label{eq:final_memory_replay}
\vspace{-2mm}
\end{equation}

Here, $M_{batch}^{final}$ and $M_{replay}^{final}$ are the final memory reservations for batch execution and replay buffer, respectively. These equations ensure that the total memory used is exactly $M$, and that it is distributed in proportion to the desired reservations for batch execution and the replay buffer.

% \todo{example begins here, may place in Appendix? }

Let's consider different scenarios to illustrate the behavior of the memory reservation system:

\begin{itemize}[leftmargin=10px]
    \item \textbf{Timing Inefficiency}: When $\alpha > 1$ (indicating an episode runtime exceeding its subtask deadline), there is an increase in the batch memory allocation ($M_{batch}^{new}$). This allocation is unaffected by performance ($\beta \geq 1$), as efficient operation negates the need for additional batch processing memory.
    \item \textbf{Performance Deterioration}: An increase in replay buffer memory allocation ($M_{replay}^{new}$) is observed when $\beta < 1$, signifying performance deterioration. Temporal efficiency ($\alpha \leq 1$) neutralizes the adjustment to batch memory ($M_{batch}^{new}$), as an efficient runtime eliminates the need for more replay buffer memory.
    \item \textbf{Maintaining Optimal Performance}: Under optimal conditions of both runtime and performance ($\alpha \leq 1$, $\beta \geq 1$), the memory reservations for batch execution and replay buffer remain constant. This indicates the system is running efficiently, thus no need to adjust memory allocations.
    \item \textbf{Balanced Trade-off}: In the event of both inefficient runtime ($\alpha > 1$) and performance deterioration ($\beta < 1$), there is a compensatory increase in both $M_{batch}^{new}$ and $M_{replay}^{new}$. Despite this, the overall memory usage must remain within system constraints, necessitating a proportional reduction in memory reservations if required.
\end{itemize}

These illustrative cases demonstrate the adaptability of the memory reservation system in response to different performance conditions. By dynamically adjusting $M_{batch}$ and $M_{replay}$, the system can address diverse challenges and maintain a balanced trade-off between timing performance and algorithm performance, all within the constraints of the available memory budget.

\subsection{Algorithm Details}

\begin{algorithm}[t]
\caption{\Approach{} framework}
\label{alg:main_method}
\begin{algorithmic}[1]
\State \textbf{Input:} hyperparameter $m$; hyperparameter 
 $\gamma$; end-to-end deadline $D$; training budget $B$; maximal training episode $N$; Data frame information $I$; training budget $C$;
% \State \textbf{function} \textsc{\Approach{}}($a$, $D$, $N$);
\State Initialize memory reservation \{$M_{batch}, M_{replay}$\} with $m$;
\State Initialize spent training cost $c$ with $0$;
\For {each episode $i \in N$} % \Comment{DRL training}
\State Batch size $b_i\leftarrow$\textsc{BatchSizeControl}$(D, B, M_{batch})$;
\State Replay size $r_i \leftarrow \textsc{ReplayControl}(I, M_{replay})$;
\If{$i > 0 \text{ and } r_i < r_{i-1}$}
    \State \textsc{ShrinkReplayBuffer($r_i$)};
    \State \textsc{GarbageCollection()};
\Else
    \State\textsc{ExpandReplayBuffer($r_i$)};
\EndIf
\State \textsc{Synchronization()};
\If{\textsc{IsCoarseGrained()}}
    \State Training Logs $s$ $\leftarrow$ \textsc{VanillaTraining()};
\Else
    \State Training Logs $s$ $\leftarrow$ \textsc{DynamicTraining}();
\EndIf

\State Update spent training cost $c \leftarrow c + s[cost]$;
\If{$c >= C$ or meet early exit condition}
    \State \textsc{exit()};
\EndIf
\State Update memory reservation \{$M_{batch}, M_{replay}$\} $\leftarrow$ \textsc{RuntimeCoordinator} $(s, \gamma, \{M_{batch}, M_{replay}\})$
\EndFor

\end{algorithmic}
\end{algorithm}

The overall workflow of the \Approach{} framework is presented in Algorithm~\ref{alg:main_method}. For each DRL training episode, the batch size is initially determined in accordance with Eq.~\ref{eq:linear_scaling} in Sec.~\ref{sec:deadline_driven_feedback}. Subsequently, the replay buffer size is computed by Eq.~\ref{eq:max_replay_buffer_size} in Sec.~\ref{sec:effcient_memory_managemnet}. Following this, a series of replay buffer size adjustments, memory garbage collection, and synchronization operations are executed on-demand to maintain system correctness.
% Next, several timestep minibatch training is conducted, adhering to the standard training process. 
Next, the coarse-grained algorithm is trained according to the standard process, while the fine-grained algorithm slightly adjusts the batch size of each training step.
Training progress is checked afterward to avoid overspending the training budget. It's worth mentioning that inspired by early stopping techniques~\cite{yao2007early}, once the training meets the algorithm performance goal($K$ consecutive episodes rewards consistently no smaller than the targeted algorithm performance $R^{*}$), the training process will exit early to minimize latency.
Lastly, based on the acquired training logs, the runtime coordinator dynamically adjusts the memory resources allocated to batch execution and replay buffer management by Eq.~\ref{eq:final_memory_replay}. This enables the \Approach{} framework to effectively balance the trade-offs among data parallelism, memory usage, and algorithm performance, resulting in a more efficient and adaptive system. 

\section{Evaluation}

% \zexin{Further, in evaluation, we need to present our solution under different-level workloads (normal, high, etc.) and high computational demand / overloaded scenarios. }

\subsection{Experimental Setup}

% \subsubsection{Testbeds}
\noindent\textbf{Testbeds.}
We choose three different platforms aiming to demonstrate the cross-platform compatibility of our design as shown in Table~\ref{tab:hardware}. 
These platforms comprise one desktop configuration along with two NVIDIA embedded platforms, motivated by the widespread adoption of NVIDIA hardware in deployed autonomous systems, particularly in the domains of autonomous driving~\cite{kato2018autoware,kisavcanin2017deep} and robotics~\cite{popov2022nvradarnet,Duckiebot(DB-J),SparkFun_JetBot,Waveshare_JetBot}.

\begin{table}[!htbp]
  \centering
  \caption{Hardware platforms used in our experiments.}
  \renewcommand\arraystretch{1.3}
  \resizebox{0.45\textwidth}{!}{ 
    \begin{tabular}{|c|c|c|c|}
    \hline 
     & {\textbf{Desktop}} & {\textbf{NVIDIA AGX Xavier}} & {\textbf{NVIDIA AGX Orin}}\\
    \hline
    {\multirow{3}{*}{CPU}}  & Intel(R) Core(TM)  & 8-core NVIDIA   & 8-core Armv8.2\\
    & i7-10700K CPU & Carmel Armv8.2  & Cortex-A78AE \\
    & @ 3.80GHz & 64-bit CPU & 64-bit CPU \\
    \hline
    GPU & NVIDIA GTX 3060 & NVIDIA Volta GPU & NVIDIA Ampere GPU \\
    \hline
    Memory & DDR4 16GBx2 & 16GB LPDDR4x & 32GB LPDDR5 \\
    \hline
    Storage & 1TB SSD & 32GB eMMC & 64GB eMMC \\
    \hline
    \end{tabular}
}
  \label{tab:hardware}
  \vspace{-0mm}
\end{table}

\noindent\textbf{Benchmarks.}
To thoroughly evaluate our solution, we have selected three deep reinforcement learning benchmark environments, as detailed in Table~\ref{tab:env}. These environments encompass various application areas and are widely used in the research community. Note that we evaluate two representative DRL benchmarks, Atari~\cite{mnih2015human} and Classic Control~\cite{6313077}, integrated with widely used autonomous-learning-library(ALL)~\cite{nota2020autonomous}. 
% Specifically, Classic Control often emphasizes the importance of control theory concepts, allowing us to validate the applicability of our approach in robotic contexts.  Moreover, we've selected Atari, which features complex, partially observable environments and requires long-term strategic planning akin to a robot navigating unstructured terrain. These complexities mimic challenges in robotic applications, such as managing partial observability, making long-term decisions, and developing efficient exploration strategies. Hence, the Atari benchmark allows us to assess our approach's efficacy in the context of more complex robotic tasks.
We use these Classic Control and Atari to evaluate the overall effectiveness and versatility of \Approach{}.
Additionally, to demonstrate the robustness of our solution in real-world practical scenarios, we further conduct a practical case study by integrating our approach into autonomous navigation scenarios, empowered by a high-resolution simulator DonkeyCar~\cite{bib:gymdonkeycar}. In our experiments, we assess the performance of three well-known and widely implemented DRL algorithms: DQN~\cite{mnih2015human}, DDQN~\cite{van2016deep}, and C51~\cite{bellemare2017distributional}. 
% \textcolor{red}{We've only tested one algorithm for donkeycar right?} 
These algorithms provide a comprehensive comparison of our solution's efficacy and adaptability across different learning methods and application domains. We set hyperparameter $b_{min}$ strictly following the MAX-A preset settings, $m$ refers to each platform's maximal available memory, $\gamma$ to 4. 
We set hyperparameter maximal episode number $N$ to 10000 for all environments, data budget the same as MAX-A preset setting, and early exit parameter $K$ to 10 for all benchmark,  $R^{*}$ to 200, 300, and 1000 for Classic Control, Atari, and DonkeyCar. 

\begin{table}[!t]
\begin{center}
\caption{Deep reinforcement learning benchmark environments used in our experiments.}
\begin{tabular}{|c|p{0.65\linewidth}|}
 \hline
Benchmark ID & \multicolumn{1}{c|}{Description} \\
\hline
\multirow{2}{*}{Atari~\cite{mnih2015human}} & A well-known environment for video game-based deep reinforcement learning research, featuring a collection of Atari 2600 games. \\
\hline
\multirow{2}{*}{Classic Control~\cite{6313077}} & A set of classical control games for deep reinforcement learning research, capable of simulating robots and complex mechanical systems. \\
\hline 
\multirow{2}{*}{DonkeyCar~\cite{bib:gymdonkeycar}} & A practical simulated deep reinforcement learning environment for training autonomous driving agents and deploying to real-world platforms. \\
 \hline
\end{tabular}
\label{tab:env}
\end{center}
\vspace{-0mm}
\end{table}

\noindent\textbf{Metrics.}
This paper evaluates two main types of metrics. The first set reflects latency predictability, which evaluates the throughput by end-to-end latency and real-time performance indicators using task deadline miss rates. Moreover, end-to-end deadlines for DRL training are based on the worst-case execution time (WCET), and task deadlines are based on the proportional intermediate deadline assignment.
The second set of metrics corresponds to algorithm performance, measuring widely adopted maximal and average cumulative rewards.

% \subsubsection{Baselines}
\noindent\textbf{Baselines.}
We compare \Approach{} to the following approaches: 
% \yufei{Can use [leftmargin=*] to remove indent for itemize}

\begin{itemize}[leftmargin=10px]
    \item \textbf{MAX-A}: To maximize the algorithm performance, we directly adopt preset hyperparameters in the autonomous learning library (ALL)~\cite{nota2020autonomous}, which is an object-oriented novel deep reinforcement learning (DRL) library designed for PyTorch~\cite{paszke2019pytorch}. The creators have incorporated high-quality reference implementations of modern DRL algorithms and provided ease of use API, and pre-defined hyperparameter sets for best algorithm performance.
    \item \textbf{MAX-P}: MAX-P, or maximal data parallelism, denotes a high-efficiency mode that enables concurrent data processing to achieve peak performance. For fair comparisons, we implement MAX-P strictly following the implementation of ALL~\cite{nota2020autonomous} to exploit the computational resources and maximize throughput in DRL training scenarios.
    \item \textbf{\Approach{}}: Our proposed solution is implemented at various DRL training granularities: \Approach{}$_{episode}$ for coarse-grained episode-level and \Approach{}$_{step}$ for fine-grained step-level implementation. 
\end{itemize}

%% DRL training jobs need to be sequential because of dependencies between consecutive tasks. In addition, the underlying hardware uses only a single unit of computation (single GPU) and thus the only scheduling policy available for us to use is FIFO.

\noindent\textbf{Scheduling Policy.}
In on-device DRL training, there is only one DRL training job running at  a time because the training task is extremely resource-demanding for both computation and memory resources, and the GPU resource is often limited (e.g., typically only one GPU device available for most desktop and embedded settings). Therefore, such scenarios cannot be optimized through multi-task scheduling. We thus applied FIFO scheduling in all cases. Deadline miss rates are calculated task-wise based on the proportional intermediate deadline assignment policy.

\subsection{Overall Effectiveness}

\label{sec:overall_effectiveness}

\begin{figure*}[!htbp]
    \centering
    \includegraphics[width=\textwidth]{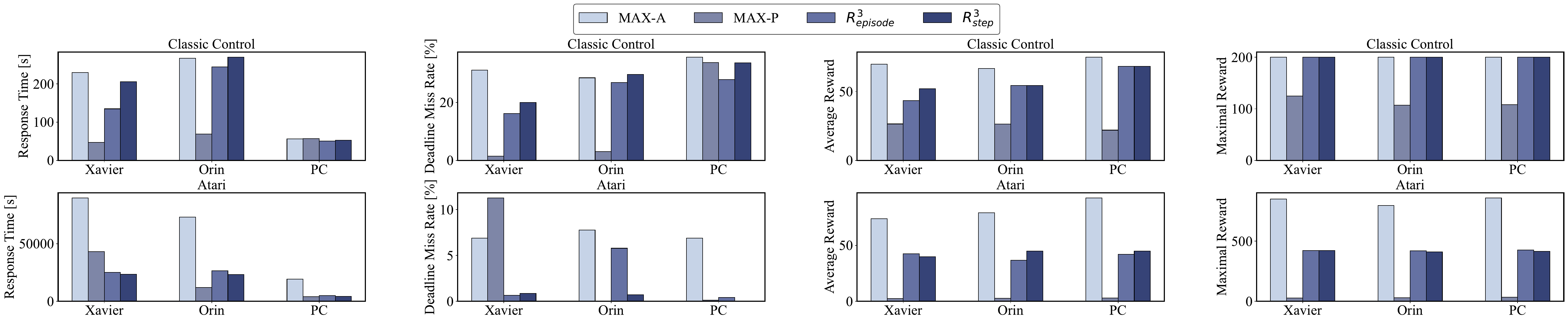}
    \caption{Overall effectiveness of \Approach{} on the C51 algorithm evaluated on four different resource-constrained intelligent robotic systems. \Approach{} auto balances throughput, timing correctness, and algorithm performance.}
    \vspace{-3mm}
    \label{fig:overall_effectiveness}
\end{figure*}

\begin{figure*}[!t]
    \centering
    \includegraphics[width=\textwidth]{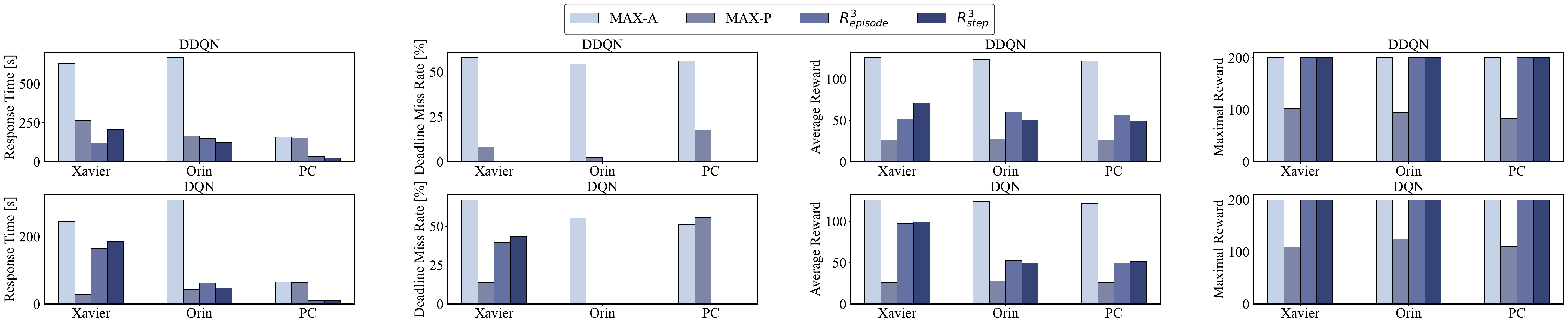}
    \caption{\Approach{} evaluated on more different DRL algorithms on the Classic Control benchmark.}
    \label{fig:overall_effectiveness_supp}
    \vspace{-3mm}
\end{figure*}

We measure the overall effectiveness of \Approach{} across three platforms. Since our design concerns latency predictability and algorithm efficacy, we measure both constraints and compare against strong baselines based on widely used DRL evaluating benchmarks autonomous-learning-library (ALL)~\cite{nota2020autonomous}. 

\begin{figure}[!t]
    \centering
    \includegraphics[width=0.45\textwidth]{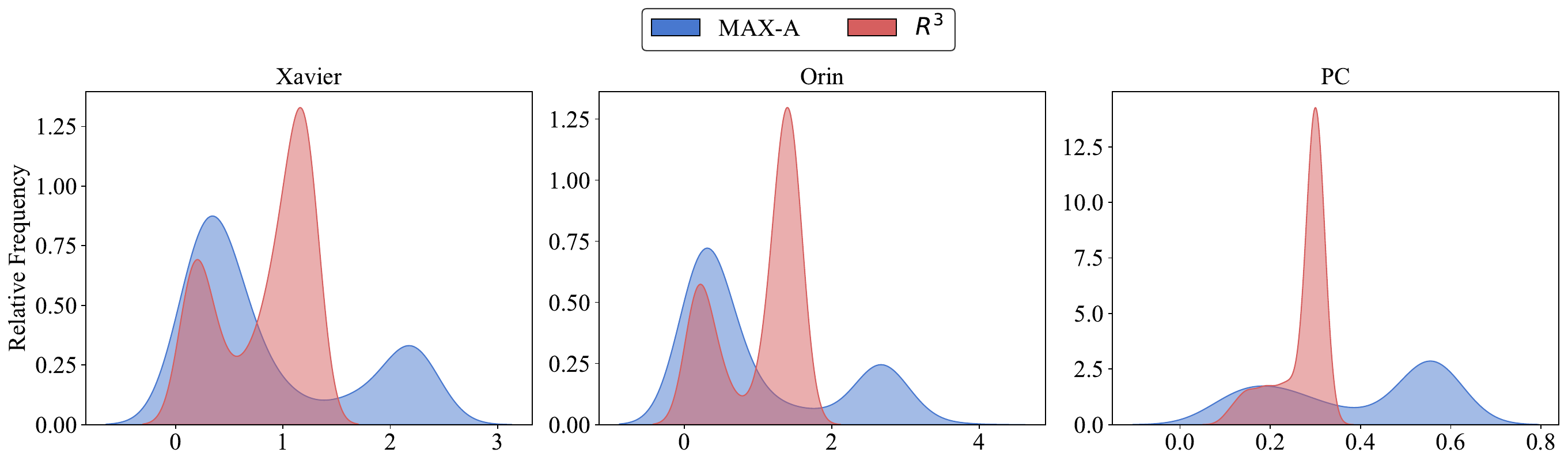}
    \caption{Episode-level response time histogram across different platforms.
    \Approach{} ensure significantly better timing correctness than MAX-A qualitatively. }
    \vspace{-3mm}
    \label{fig:hist}
\end{figure}

\noindent \textbf{Latency predictability.}
Figure~\ref{fig:overall_effectiveness} shows timing performance comparisons. In the Classic Control benchmark, \Approach{} outperforms MAX-A, with average execution time improvements of 39.3\%, 8.2\%, and 9.0\% on Xavier, Orin, and PC. This improvement is attributed to our memory management, which increases memory access slightly but enhances timing optimization. On the PC, latencies are similar, maybe due to the high computation capabilities of desktop GPU meeting the relatively small benchmark Classic Control. Other platforms show notable timing differences, indicating the efficacy of our solution on embedded systems. The deadline miss rate further emphasizes this, with \Approach{} improving over MAX-A by 15.0\%, 1.7\%, and 7.8\% on Xavier, Orin, and PC. On the Atari benchmark, \Approach{} balances timing correctness and algorithm performance, outperforming MAX-A by 71.9\% and 5.8\%. However, MAX-P on Xavier is slower, possibly due to the Atari benchmark's longer duration and Xavier's older Linux kernel affecting CPU scheduling. Figure~\ref{fig:hist} breaks down response times, showing \Approach{} consistently outperforms MAX-A in timeliness. The empirical results demonstrate that our proposed \Approach{} qualitatively has significantly better timeliness than MAX-A, avoiding potentially high-tailed latency.

\begin{figure}[!t]
    \centering
    \includegraphics[width=0.47\textwidth]{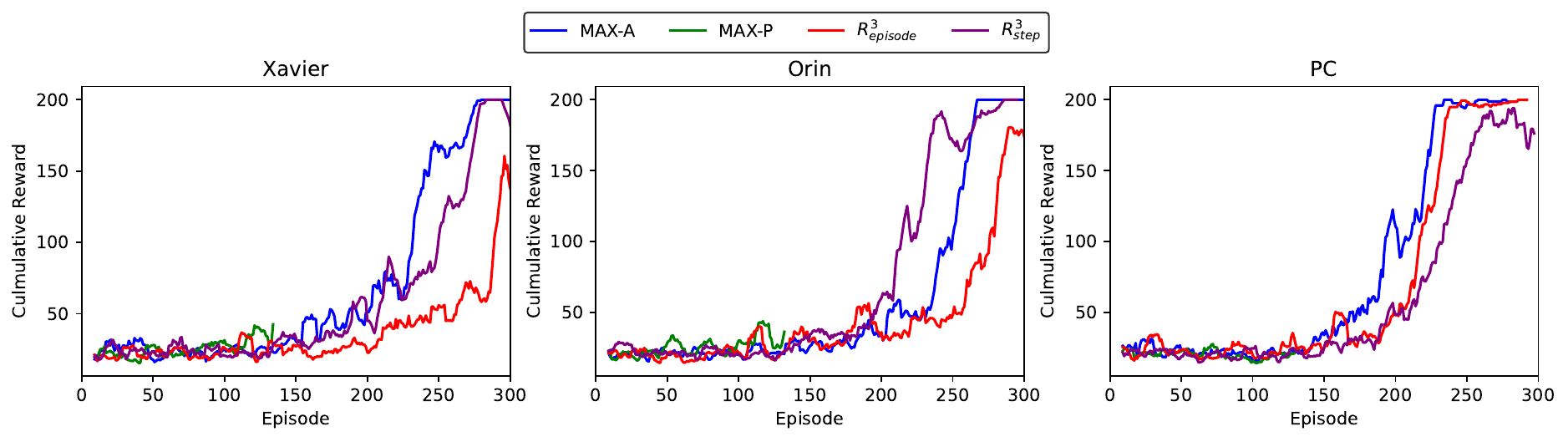}
    \caption{
    Episode-level reward curves across different platforms. % on Classic Control benchmark. 
    The algorithm performance of \Approach{} is consistently better than MAX-P and close to MAX-A (empirical optimal).}
    \vspace{-3mm}
    \label{fig:per-episode_rewards}
\end{figure}

\noindent \textbf{Algorithm Efficacy.} The last two columns of Figure~\ref{fig:overall_effectiveness} display algorithm performance comparisons. On the Classic Control benchmark, \Approach{} almost obtains a very close to optimal algorithm performance (MAX-A) by on average 100.0\% and 82.7\% on average rewards and maximal rewards, while outperforming MAX-P by a large margin on average by 76.5\% and 133.6\% on average rewards and maximal rewards, respectively. Note that the theoretical upper bound of maximal reward for Classic Control is 200; both MAX-A and \Approach{} consistently reach this maximal value during training. This quantitatively validates the effectiveness of our co-optimization, especially for training a usable DRL algorithm under strict time and memory constraints. On the Atari benchmark, \Approach{} auto-balances the timing correctness and algorithm performance. Specifically, \Approach{} achieves on average by 49.7\% and 50.2\%  on the average rewards and maximal rewards than MAX-A, but largely outperforms on average by 1521.1\% and 1297.8\% on the average rewards and maximal rewards than MAX-P. This supports the validity \Approach{} and further evidences the resilience of our design.
Furthermore, to demonstrate quantitatively the algorithm performance between different methods, 
we plot the per-episode cumulative rewards during the training process, as shown in Figure~\ref{fig:per-episode_rewards}. It can be observed that MAX-P runs out of data budget too early (green curves), so the training lasts only a very small number of episodes and therefore leads to very poor algorithm performance. In contrast, our proposed \Approach{} selects proper bath size in the batch execution, thus reaching algorithm performance consistently better than MAX-P and nearly approaching empirically optimal MAX-A.

\noindent \textbf{Evaluated on more DRL algorithms.} 
Our method, \Approach{}, was further evaluated on two additional algorithms (refer to Figure~\ref{fig:overall_effectiveness_supp}). Specifically, for the DDQN algorithm, \Approach{} performs better than MAX-A for execution time by 73.9\%, 79.4\%, and 81.1\% on Xavier, Orin, and PC. Also, \Approach{} performs better than MAX-P for maximal rewards by 94.2\%, 110.5\%, and 140.1\% on Xavier, Orin, and PC. Additionally, \Approach{} performs better than MAX-P for average rewards by 94.2\%, 133.1\%, and 141.0\% on Xavier, Orin, PC.
Additionally, for the DQN algorithm, \Approach{} performs better than MAX-A for execution time by 133.1\%, 102.0\%, and 101.3\% on Xavier, Orin, and PC. Also, \Approach{} performs better than MAX-P for maximal rewards by 83.5\%, 83.8\%, and 60.0\% on Xavier, Orin, and PC. Additionally, \Approach{} performs better than MAX-P for average rewards by 256.8\%, 84.3\%, and 68.3\% on Xavier, Orin, and PC. Our efficient memory optimization yields supreme latency performance of \Approach{}, even outperforming MAX-P in DDQN and DQN. Note that \Approach{} exhibited a markedly low deadline miss rate for both DRL algorithms, suggesting its adaptability and efficacy in enhancing real-time performance (see Table~\ref{tab:deadline}).

\begin{table}[!t]
\centering
\caption{Deadline miss rate on DDQN and DQN algorithms. The best values are in bold.}
\resizebox{0.35\textwidth}{!}{ 
\begin{tabular}{c|c|ccc}
\hline
& Solution & Xavier & Orin & PC \\
\hline
{\multirow{4}{*}{DDQN}} & MAX-P & 8.3\% & 2.4\% & 17.7\% \\
& MAX-A & 57.9\% & 54.5\% & 56.0\% \\
& \Approach{}$_{episode}$ & \textbf{0.0\%} & \textbf{0.0\%} & \textbf{0.0\%} \\
& \Approach{}$_{step}$ & \textbf{0.0\%} & \textbf{0.0\%} & \textbf{0.0\%} \\
\hline
{\multirow{4}{*}{DQN}} & MAX-P & \textbf{13.8\%} & \textbf{0.0\%} & 55.6\% \\
& MAX-A & 67.1\% & 55.4\% & 51.4\% \\
& \Approach{}$_{episode}$ & 39.6\% & \textbf{0.0\%} & \textbf{0.0\%} \\
& \Approach{}$_{step}$ & 43.6\% & \textbf{0.0\%} & \textbf{0.0\%} \\
\hline
\end{tabular}}
\label{tab:deadline}
\vspace{-3mm}
\end{table}

\begin{minipage}{0.45\textwidth}
\begin{shaded}
    \noindent{\textbf{Cross-platform overall effectiveness}}: \Approach{}
    effectively addresses all challenges outlined in Sec.~\ref{sec:case_study} for resource-constrained intelligent robotic systems. It auto-balances throughput, timing correctness, and algorithm performance in DRL training across different platforms.
\end{shaded}
\end{minipage}

% \subsection{Breakdown Analysis: Why \Approach{} works? \zexin{Make it optional?? @Cong}}

% \label{sec:breakdown}

% \todo{low priority: 1. not claimed in contribution 2. there exists one breakdown analysis in car case study}

% BaseGPU, BaseGPU + DFL (deadline-driven feedback loop), BaseGPU + EMM (efficient memory mangement), \Approach{} w/o Coord, \Approach{}

% evaluation flow 1: dummy baseline, baseline + dynamic batch, baseline + memory mangement, ours
% evaluation flow 2: no coordination has some issues. compare with w/o coordination

\subsection{A Practical Case study: Autonomous Navigation via DRL}

\label{sec:car}

\begin{figure}[!t]
    \centering
    \begin{subfigure}[b]{0.220\textwidth}\includegraphics[width=\textwidth]{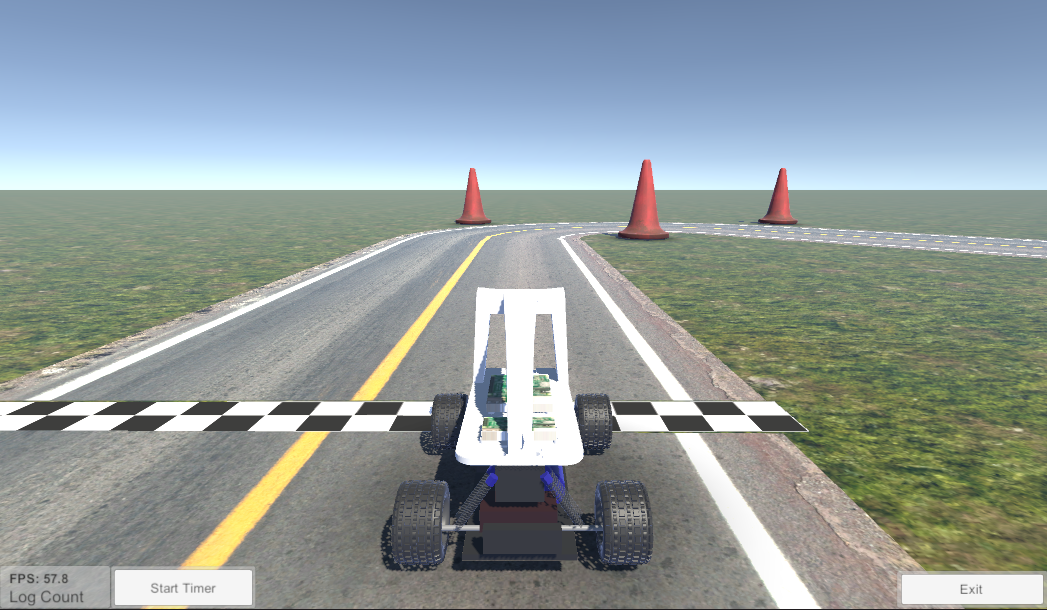}
    \caption{Generated track}
    \label{fig:generated_track}
    \end{subfigure}
    \centering
    \begin{subfigure}[b]{0.220\textwidth}\includegraphics[width=\textwidth]{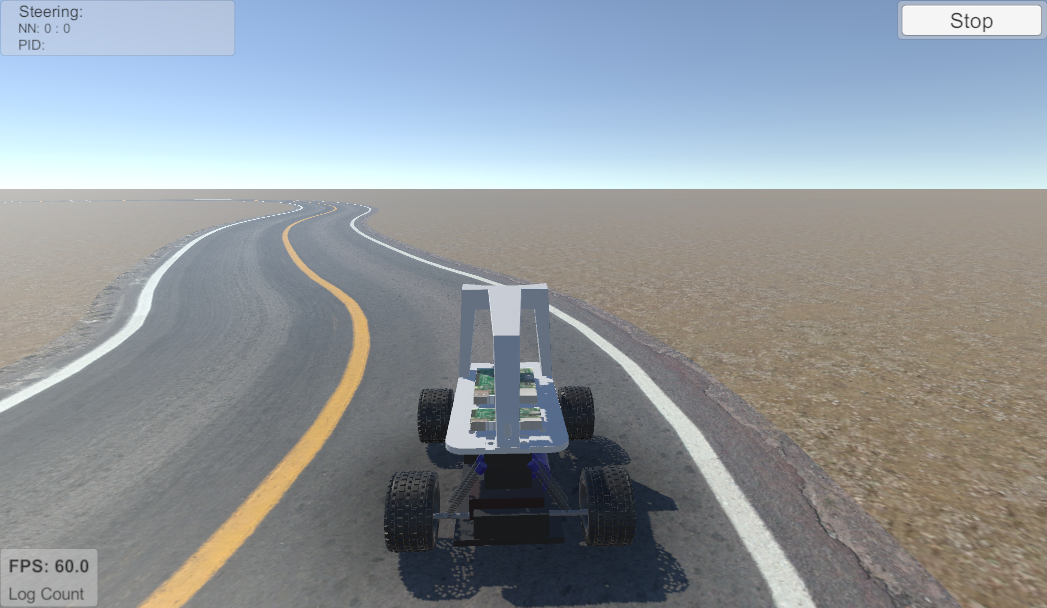} 
    \caption{Generated road}
    \label{fig:generated_road}
    \end{subfigure}
    \centering
    \begin{subfigure}[b]{0.220\textwidth}\includegraphics[width=\textwidth]{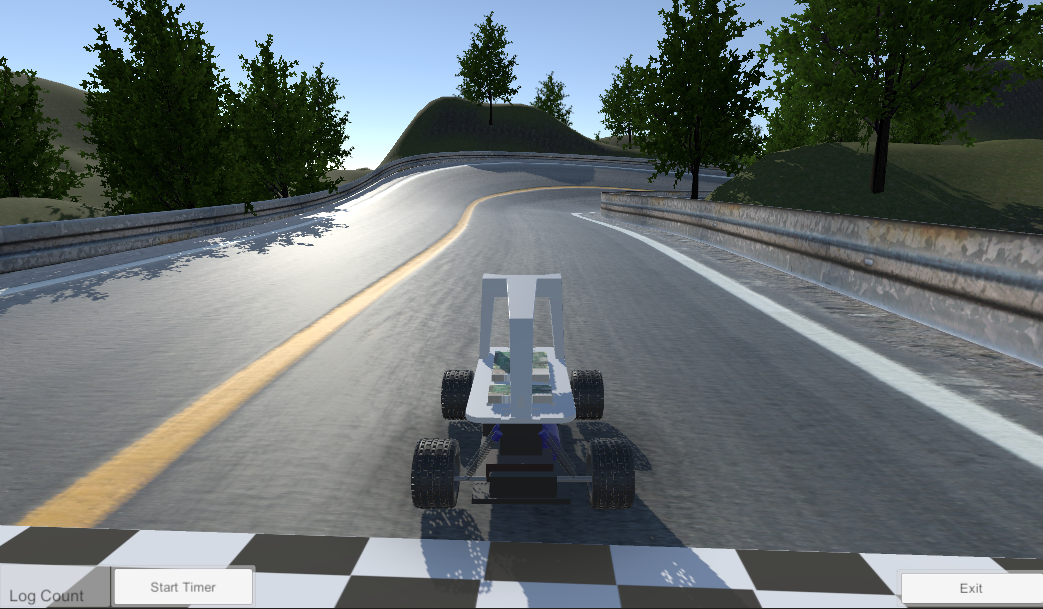}
    \caption{Mountain track}
    \label{fig:mountain_track}
    \end{subfigure}
    \centering
    \begin{subfigure}[b]{0.220\textwidth}\includegraphics[width=\textwidth]{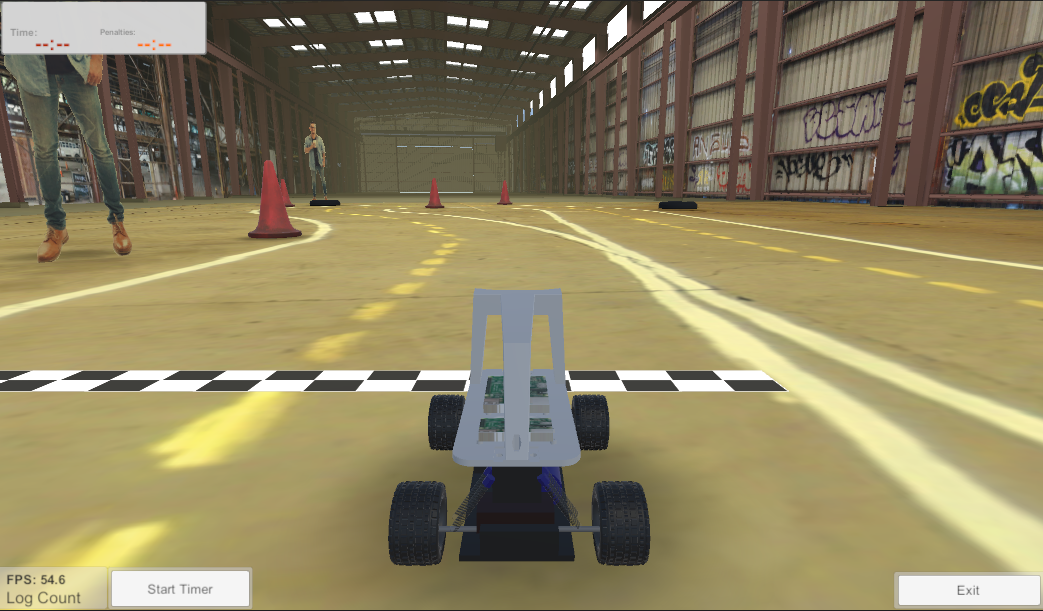}
    \caption{Warehouse track}
    \label{fig:warehouse_track}
    \end{subfigure}
    \caption{DonkeyCar~\cite{bib:gymdonkeycar} driven high-resolution autonomous navigation simulator for evaluating DRL algorithm.}
    \vspace{-2mm}
    \label{fig:donkeycar}
\end{figure}

To empirically evaluate the effectiveness of our proposed approach, \Approach{}, in real-world DRL applications for robotics, we conducted a comprehensive case study centered on autonomous navigation. We selected DonkeyCar~\cite{bib:gymdonkeycar} as the platform for this case study, a high-resolution environment based on the x86 Unity3D autonomous navigation simulator~\cite{haas2014history}, as displayed in Figure~\ref{fig:donkeycar}.
DonkeyCar was chosen due to its relevance in investigating the deployment of DRL-based control systems in real-world robotics contexts. Several real-world autonomous driving cars are built upon DonkeyCar library~\cite{bib:donkeycar_build,bib:donkeycar_s1} based on TensorFlow framework~\cite{tensorflow}.
To successfully incorporate DRL training into this platform, we implement a highly optimized networking module that facilitates the seamless execution of DRL training natively on advanced embedded systems (Xavier and Orin).\footnote{Due to specific compilation challenges, we were unable to prepare the DRL training library for the x64 PC environment. Table~\ref{tab:donkey_car} does not include the evaluation of DonkeyCar on the PC.} The integration of these robust systems underscores the real-world feasibility and performance of \Approach{} in robotic contexts, highlighting its potential for wider adoption in various robotic applications. 

DonkeyCar~\cite{bib:gymdonkeycar} processes 60 FPS high-resolution RGB images as streaming camera inputs, raising a significant memory challenge when conducting on-device DRL training. Table~\ref{tab:donkey_car} presents the results of the DDQN algorithm implementation on DonkeyCar, showcasing a range of evaluation metrics. Specifically, we report latency, FPS to describe throughput, data budget consumption percentage to describe the relative percentage of training completion, and maximal rewards R$_{max}$ and average rewards R$_{avg}$ to describe algorithm performance.

\begin{table}[!t]
\centering
\caption{Quantitative results on DonkeyCar simulator. DPR implies data processing rate, Reward$_{max}$ implies maximal episode rewards, and Reward$_{avg}$ implies average episode rewards. {$^*$} implies OOM occurs. }
\resizebox{0.48\textwidth}{!}{ 
\begin{tabular}{c|c|ccccc}
\hline
Platform & Solution & Latency$(\downarrow)$ & DPR$(\uparrow)$ & Budget$(\uparrow)$ & R$_{max}$$(\uparrow)$ & R$_{avg}$$(\uparrow)$ \\
\hline
\hline
\multirow{4}{*}{Xavier} & MAX-P & 64.7\textcolor{black}{$^*$} & 1423.8  & 0.2\% & 116.6 & 59.8 \\
 & MAX-A & 3619.4\textcolor{black}{$^*$} & 742.0 & 41.9\% & 1343.2 & 157.3 \\
 & \Approach{}$_{episode}$ & 7525.9 & 850.4 & 100.0\% & 532.7 & 79.1 \\
 & \Approach{}$_{step}$ & 7429.1 & 861.5 & 100.0\% & 544.1 & 80.2 \\
\hline
\multirow{4}{*}{Orin} & MAX-P & 4918.0 & 1301.3 & 100.0\% & 29.4 & 7.3 \\
 & MAX-A & 8341.0\textcolor{black}{$^*$} & 653.7 & 85.2\% & 1303.8 & 156.9 \\
 & \Approach{}$_{episode}$ & 8416.18 & 760.4  & 100.0\% & 598.6 & 67.0 \\
 & \Approach{}$_{step}$ & 8205.23 & 780.0  & 100.0\%& 588.5 & 70.2\\
\hline
\end{tabular}}
\label{tab:donkey_car}
\vspace{-3mm}
\end{table}

Results demonstrate that \Approach{} adeptly balances both latency predictability and algorithm efficacy without incurring out-of-memory (OOM) problems on both platforms; while OOM occurs for MAX-A on both platforms and MAX-P on Xavier.
\Approach{}'s data processing rate (DPR) exceeds MAX-A by an average of 16.6\%, while \Approach{} lags behind MAX-P by an average of 40.4\%. This is because \Approach{} provides a smaller batch size than MAX-P.
In terms of algorithm performance, \Approach{} significantly outperforms MAX-P, with maximal and average rewards better by 872.9\% and 1140.3\% in the challenging autonomous navigation task. Still, \Approach{} trades off certain algorithm performance by 57.2\% and 52.8\% on maximal and average rewards, mainly due to reserving smaller memory for replay buffer compared to MAX-A.

\begin{figure}[!t]
\centering
\includegraphics[width=0.5\textwidth]{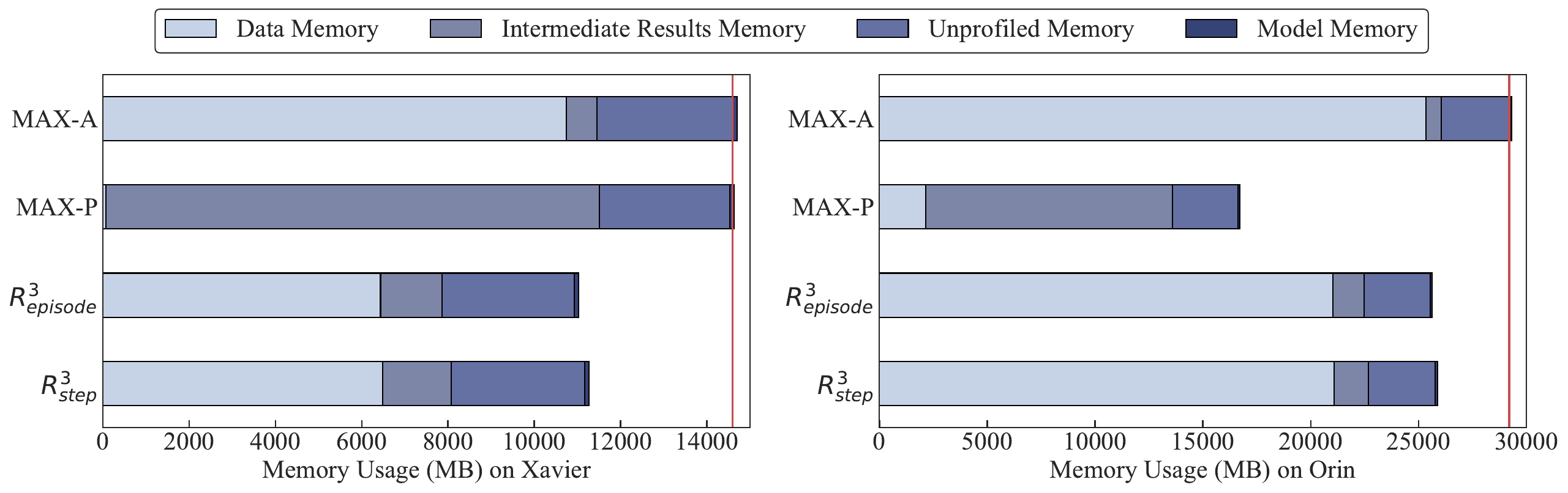}
\caption{Memory breakdown on DonkeyCar evaluated on two embedded devices. The red line represents the OOM point.} 
\vspace{-4mm}
\label{fig:memory_breakdown_donkeycar}
\end{figure}

A deeper dive into \Approach{}'s memory efficiency, as shown in Figure~\ref{fig:memory_breakdown_donkeycar}, reveals its strengths. MAX-A heavily consumes memory in data storage due to high-res RGB images in its replay buffer. As this buffer expands, memory use grows uncontrollably, causing OOM errors. In contrast, MAX-P's memory consumption spikes in batch execution, storing numerous intermediate activations, leading to OOM errors on Xavier. However, \Approach{} optimally manages memory in both data storage and batch execution. Its innovative replay buffer design allows for a larger size within the same memory constraints, enhancing algorithmic output. \Approach{} also dynamically adjusts batch sizes during training, striking a balance between timing requirements and memory constraints, ensuring effective DRL training within these boundaries.

\begin{minipage}{0.45\textwidth}
\begin{shaded}
    \noindent{\textbf{Pratical usability}}: This case study demonstrates the integration effectiveness of \Approach{} on practical robotic applications. \Approach{} significantly reduces the memory footprint by efficient memory optimization, ensuring the usability of \Approach{} in embedded robotic DRL training scenarios.
\end{shaded}
\end{minipage}

\subsection{Overhead Analysis} 
\label{sec:overhead}
We further break down the overhead of \Approach{} for different platforms regarding execution time and memory usage.

\begin{table}[!htbp]
    \centering
    \renewcommand\arraystretch{1.0}
    \caption{Detailed average runtime execution overhead (s) and percentage of applying \Approach{} for each benchmark.} 
    \resizebox{0.48\textwidth}{!}{ 
    \begin{tabular}{c|c|ccc}
    \hline
    % & \multicolumn{3}{c|}{Xavier} & \multicolumn{3}{c|}{Orin} & \multicolumn{3}{c}{PC}\\
    % \hline
    Platform & Benchmark & Feedback & MM & Coord \\
    \hline
    {\multirow{3}{*}{Xavier}} & Classic Control & 0.30 (0.15\%) & 0.11 (0.05\%) & 0.07 (0.04\%) \\
    & Atari & 3.68 (0.01\%) & 10.22 (0.01\%) & 0.16 (0.01\%)  \\
    & DonkeyCar & 0.35 (0.01\%) & 41.95 (0.56\%) & 0.09 (0.01\%) \\
    \hline
    {\multirow{3}{*}{Orin}} & Classic Control & 0.14 (0.05\%) & 0.29 (0.11\%) & 0.06 (0.02\%) \\
    & Atari & 1.55 (0.03\%) & 11.04 (0.02\%) & 0.08 (0.01\%)  \\
    & DonkeyCar & 0.22 (0.01\%) & 115.8 (1.41\%) & 0.04 (0.01\%) \\
    \hline
    {\multirow{3}{*}{PC}} & Classic Control & 0.05 (0.09\%) & 0.10 (0.19\%) & 0.01 (0.02\%) \\
    & Atari & 0.99 (0.01\%) & 8.93 (0.02\%) & 0.04 (0.01\%)  \\
    & DonkeyCar & N/A & N/A & N/A \\
    \hline
    \end{tabular}
    }
    \label{tab:execution_overhead}
    \vspace{-3mm}
\end{table}

\noindent \textbf{Execution Overhead.}
Table~\ref{tab:execution_overhead} details the average execution overhead for each component: deadline-driven feedback loop (Feedback), replay memory management (MM), and runtime coordinator (Coord) across platforms. The system's execution overhead remains under 1.43\%, underscoring our method's efficiency. The feedback loop has a minimal overhead, peaking at 3.68s in Atari. The runtime coordinator also maintains a low overhead, at most 0.16s and 0.01\%. Conversely, memory management has a higher overhead, reaching 115.8s and 1.41\% on Orin, due to increased memory access in our strategy. Yet, this is counterbalanced by memory optimization techniques like non-blocking data prefetching, enhancing the overall system performance.

\begin{table}[!htbp]
    \centering
    \renewcommand\arraystretch{1.0}
    \caption{Detailed memory overhead and percentage of applying \Approach{} for each benchmark.}
    \resizebox{0.48\textwidth}{!}{ 
    \begin{tabular}{c|ccc|ccc}
    \hline
    & \multicolumn{3}{c|}{(a) Overhead upon Framework} & \multicolumn{3}{c}{(b) Overall Overhead Ratio}\\
    \hline
    & Feedback & MM & Coord & Xavier & Orin & PC \\
    \hline
    Classic Control & 5 KB & 160 KB & 3 KB & 0.001\% & 0.001\% & 0.001\% \\
    Atari & 50 KB & 15.6 MB & 10 KB & 0.975\% & 0.488\% & 0.488\% \\ 
    DonkeyCar & 12 KB & 1.6 MB & 15 KB & 0.010\% & 0.005\% & 0.005\% \\
    \hline
    \end{tabular}}
    \label{tab:memory_overhead}
    \vspace{-3mm}
\end{table}

\noindent \textbf{Memory Overhead.}
Table~\ref{tab:memory_overhead} showcases the memory overhead of \Approach{}, both in absolute terms and as a percentage. The overhead remains below 1\% on all platforms, making \Approach{} ideal for memory-limited autonomous systems. While emphasizing memory efficiency, \Approach{} may compromise latency. Future work might focus on minimizing memory access or enhancing caching strategies for better memory management.

\begin{minipage}{0.45\textwidth}
\begin{shaded}
    \noindent{\textbf{Low overhead}}: \Approach{}'s efficient implementation introduces low overhead, making it a lightweight deployment solution for optimizing on-device DRL training.
\end{shaded}
\end{minipage}

\subsection{Adaptability to Different System Scenarios}

\label{sec:interference_adaptability}

As a complement to the overall effectiveness, we investigate the adaptability of the proposed \Approach{}. We focus on the adaptation of both variable deadline and variable computational interference in the following angles.

\begin{figure}[!htbp]
\centering
\includegraphics[width=0.5\textwidth]{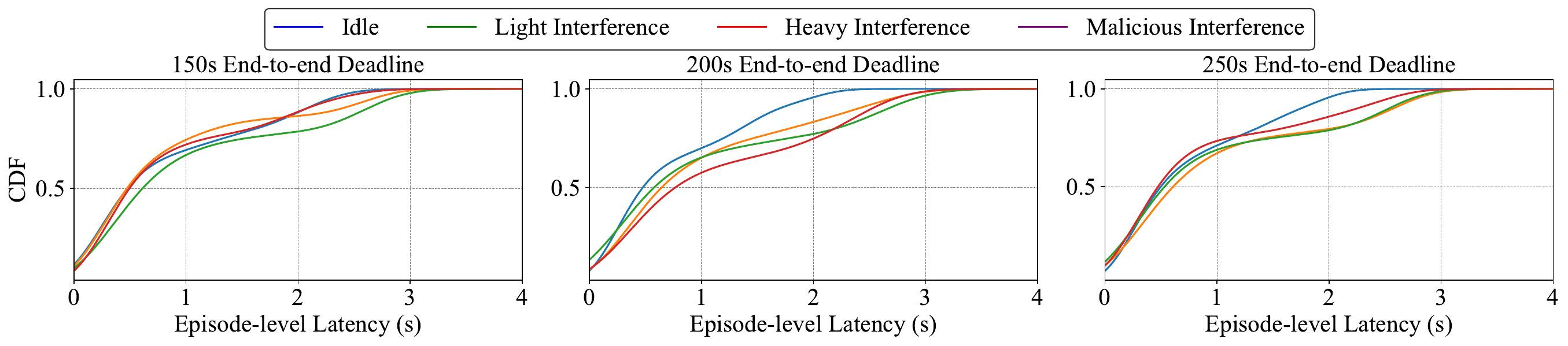}
\caption{Episode-level latency CDF for varying deadline configuration and interference workloads intensity on Xavier.}
\label{fig:interference}
\vspace{-3mm}
\end{figure}

\noindent\textbf{Adaptability to Variant Deadlines. } We examine the adaptability of \Approach{} to variant end-to-end deadlines in Figure~\ref{fig:interference}. 
This figure depicts the outcomes, demonstrating that \Approach{} can dynamically adjust to variant end-to-end deadlines. Thus it can  sustain a highly predictable task-level tail latency within a defined range of computational interference.
This evidences the resilience of \Approach{} as a solution to different timing constraints.

\noindent\textbf{Adaptability to Variant Computational Interference. }
We examine DRL training based on \Approach{} alongside a computationally demanding background workload, namely FFmpeg video decoding, to explore the adaptability of \Approach{} to computational interference. We varied the interference workload intensity by modifying the video's output resolution. Specifically, 720P, 1080P, and 2K decoding resolutions were employed to represent light, heavy, and malicious interference. Figure~\ref{fig:interference} depicts the outcomes, demonstrating that \Approach{} can dynamically adjust to computational workload, thereby sustaining a highly predictable task-level tail latency within a defined range of computational interference. Thus, it ensures the fulfillment of end-to-end deadlines.

\begin{minipage}{0.45\textwidth}
\begin{shaded}
    \noindent \textbf{Versatility}: \Approach{} consistently maintains its effectiveness across different system scenarios, including various deadline settings and interference workloads. This adaptability ensures robustness and resilience, regardless of timing or computational constraints.
\end{shaded}
\end{minipage}

\section{Related Work and Discussion}

Deep Reinforcement Learning (DRL) has gained significant attention due to its ability to learn complex tasks and adapt to dynamic environments, making it particularly suitable for recent robotics applications \cite{guo2022backdoor,peng2018deepmimic,kahn2018self,he2023robust,nikkhoo2023pimbot} and future robots with complex deep learning techniques~\cite{serengil2020lightface,li2023mimonet,afarin2023commongraph,chen2021revisiting,chen2022generate,gao2022automatic,gao2023polyscriber,chen2022nicgslowdown,chen2022learning,chen2022deepperform,li2023sibling,chen2023dynamic}.
The foundation of DRL rests on DQN's introduction~\cite{mnih2015human} and its subsequent extensions tailored to robotics, including DDQN\cite{van2016deep}, C51~\cite{bellemare2017distributional}, and others. While newer algorithms such as PPO~\cite{schulman2017proximal}, TRPO~\cite{schulman2015trust}, SAC~\cite{haarnoja2018soft} have emerged, the real-time capabilities of DRL algorithms largely remain an untapped domain.
Notably, while a few studies propose new DRL algorithms that can provide faster response times~\cite{thodoroff2022benchmarking, rtrl, garcia2015comprehensive}, \Approach{} could optimize performance of given DRL algorithms without modifying the algorithms themselves. Recent research has explored dynamic adjustments of training parameters, such as batch size~\cite{nikulin2022q} and replay buffer size~\cite{fedus2020revisiting}, to enhance algorithmic performance. However, these approaches, being agnostic to system constraints, risk causing catastrophic out-of-memory (OOM) errors as illustrated in Figure~\ref{fig:R3_challenge_overview}. In contrast, by considering the unique properties of system constraints and DRL, \Approach{}  effectively avoids OOM errors while achieving the goal of efficient real-time DRL training.

Recently, real-time ROS schedulers have emerged equipped with a suite of optimization strategies~\cite{DBLP:conf/rtss/JiangJGLTW22,DBLP:conf/rtss/LiGJGDL22,DBLP:conf/rtss/TeperGUBC22,DBLP:conf/rtss/BlassCBB21,DBLP:conf/rtss/TangFG0LD020,DBLP:conf/rtas/ChoiXK21,DBLP:conf/rtas/BlassHLZB21}. Our emphasis remains mainly on application-layer application for DRL training, but the potential of integrating more ROS optimizations remains an exciting prospect for future exploration.

Although considerable progress has been made in addressing real-time deep learning inference~\cite{neuos, kang2021lalarand, xiang2019pipelined,bateni2018apnet, zhou2018s, predjoule, DBLP:conf/rtss/NigadeBB022, DBLP:conf/rtss/JiYKASDK22, DBLP:conf/rtss/JiangLHHWXW21,jeong2022band}, unfortunately, adapting these techniques directly to the DRL training scenario poses new challenges since the complexity and resource-demanding characteristics of training. This paper empirically explores these challenges and introduces a novel framework for on-device DRL training.

\vspace{1mm}
\noindent\textbf{Limitations of \Approach{}.} \Approach{} provides optimization in a static hardware context, but there still exists space to get performance gains from software-hardware synergy~\cite{zhang2023bp}. Moreover, \Approach{} focus is on DRL training that utilizes replay buffers, leaving the optimization for replay-buffer-free DRL algorithms~\cite{schaul2016} unaddressed. Additionally, despite validations on the DonkeyCar simulator \cite{bib:gymdonkeycar} for  \Approach{}, physical-world implementations may introduce additional unpredictability, such as sensor noise, dynamic lighting conditions, unpredictable environmental changes, and unmodeled physical interactions. To understand how well \Approach{} works in the real world, a promising future direction is actual autonomous systems integration\cite{he2022robust,he2023robustev,gog2022d3,favaro2018autonomous,ap}. 

\section{Conclusion}

This study introduces the novel framework, \Approach{}, which is specifically designed to ensure timing predictability for executing deep reinforcement learning (DRL) training workloads in GPU-enabled autonomous embedded systems. The comprehensive nature of \Approach{} allows for the seamless optimization of both timing and algorithm performance while simultaneously adhering to stringent memory constraints. This is achieved by incorporating a thorough understanding of DRL workload characteristics as well as utilizing real-time system feedback information. Through rigorous experimentation, we have demonstrated that \Approach{} consistently showed a significant reduction in OOM errors, effective latency handling, and maintenance of competitive algorithm performance, marking a stride in the realm of embedded real-time DRL.

\section*{Acknowledgments}
This research was supported by the National Science Foundation under Grants CNS Career 2230968, CPS 2230969, CNS 2300525, CNS 2343653, CNS 2312397. 

\bibliographystyle{IEEEtran}
\bibliography{r3}

\end{document}